\documentclass[acmlarge]{acmart}
\usepackage[utf8]{inputenc}
\usepackage{color}
\usepackage{soul}
\sethlcolor{white}
\usepackage{rotating}
\usepackage{arydshln}
\usepackage{blindtext}
\usepackage{float}
\usepackage{stackengine} 
\stackMath
\usepackage{enumitem}

\usepackage{amsmath}

\usepackage{newunicodechar}
\newunicodechar{ℝ}{\mathbb{R}} 

\usepackage{bm} 

\AtBeginDocument{%
  \providecommand\BibTeX{{%
    \normalfont B\kern-0.5em{\scshape i\kern-0.25em b}\kern-0.8em\TeX}}}

\setcopyright{acmcopyright}
\copyrightyear{2021}
\acmYear{2021}
\acmDOI{10.1145/3503044}

\acmJournal{CSUR}
\acmVolume{1}
\acmNumber{1}
\acmArticle{1}
\acmMonth{11}

\begin{document}

\title{A Survey on Aspect-Based Sentiment Classification}

\author{Gianni Brauwers}
\email{gianni.brauwers@gmail.com}
\author{Flavius Frasincar}
\email{frasincar@ese.eur.nl}
\affiliation{%
  \institution{Erasmus University Rotterdam}
  \department{Erasmus School of Economics}
  \country{The Netherlands}
  \city{Rotterdam}
  \streetaddress{P.O. Box 1738}
  \postcode{3062 PA}
}

\renewcommand{\shortauthors}{Brauwers and Frasincar}

\begin{abstract}
  With the constantly growing number of reviews and other sentiment-bearing texts on the Web, the demand for automatic sentiment analysis algorithms continues to expand. Aspect-based sentiment classification (ABSC) allows for the automatic extraction of highly fine-grained sentiment information from text documents or sentences. In this survey, the rapidly evolving state of the research on ABSC is reviewed. A novel taxonomy is proposed that categorizes the ABSC models into three major categories: knowledge-based, machine learning, and hybrid models. This taxonomy is accompanied with summarizing overviews of the reported model performances, and both technical and intuitive explanations of the various ABSC models. State-of-the-art ABSC models are discussed, such as models based on the transformer model, and hybrid deep learning models that incorporate knowledge bases. Additionally, various techniques for representing the model inputs and evaluating the model outputs are reviewed. Furthermore, trends in the research on ABSC are identified and a discussion is provided on the ways in which the field of ABSC can be advanced in the future.
\end{abstract}



\maketitle

\section{Introduction}

The World Wide Web has provided a means for people to share their opinions online through various text-based channels, such as online reviews and social media. Being able to use these opinions to accurately assess what people think of a certain product, person, or place is highly valuable in many industries. Restaurants can adjust their menus based on online food reviews, companies can improve their products based on what consumers exactly want, and the effects of political campaigns can be evaluated by analysing social media posts. As such, with the rise of the Internet, the task of sentiment analysis has become more and more significant \cite{pang2008opinion}. 

Sentiment analysis is the task of extracting and analysing people's sentiments towards certain entities from text documents \cite{liu2012sentiment}. Sentiment analysis is sometimes also referred to as opinion mining in the literature. Yet, a distinction must be made between ``sentiments" and ``opinions". Namely, opinions indicate a person's views on a specific matter, while sentiment indicates a person's feelings towards something. Yet, the two concepts are highly related and opinion words can typically be used to extract sentiments \cite{hu2004mining, 10.5555/1597148.1597269}. In sentiment analysis, the goal is to assign sentiment polarities based on a body of text. Although the words ``polarity" and ``sentiment" are often used interchangeably, a distinction can be made again. Namely, a sentiment indicates a feeling, while the polarity expresses the orientation (e.g., \textit{``positive"}, \textit{``neutral"}, or \textit{``negative"}). The granularity of a sentiment analysis task can be described using three separate characteristics: the sentiment type, the level of the task, and the target. First, the type of sentiment output must be defined. Namely, the task can entail a simple binary classification (\textit{``positive"} or \textit{``negative"} labels), but there could also be additional labels (\textit{``neutral"}), or even a sentiment intensity or score that must be predicted. Next, the level of the task determines at which level of the text sentiment is extracted. For example, in document-level sentiment analysis, the sentiment is analysed for the entire document. For sentence-level sentiment analysis, the sentiment is analysed for each sentence within the document. There are many other levels, such as words, paragraphs, groups of sentences, or chunks of text. Lastly, the target of the task determines the focus of the sentiment. First, there may be no sentiment target, which means that the task entails assigning a sentiment score or label to the text itself. On the other hand, one may want to know the sentiment regarding specific topics, entities, or aspects within the text. To illustrate, suppose we analyse a product review. If we do not define a target, we simply consider the sentiment of the entire text. Yet, knowing exactly which aspects of the product the consumer is satisfied or dissatisfied with can be highly useful, as it provides detailed information necessary for many applications \cite{laskari2016aspect}. The task of identifying aspects and analysing their sentiments in texts is known as aspect-based sentiment analysis (ABSA). 

ABSA is a relatively new field of research that has gained significant popularity due to the many useful applications \cite{6468032}. However, ABSA and sentiment analysis generally are difficult tasks due to the general structures of sentences and writing styles \cite{schouten2015survey}. In \cite{nazir2020issues}, a comprehensive overview is provided of the issues and challenges present in ABSA. Like general sentiment analysis tasks, ABSA can be performed at multiple levels of which the document level and sentence level are typically the most popular \cite{pontiki2016semeval}. Document-level ABSA methods are concerned with finding general aspects linked to a certain entity in the document and assigning sentiments to them. Sentence-level ABSA approaches, on the other hand, attempt to identify all aspects for each sentence individually, determine the sentiment associated with these aspects, and possibly aggregate sentiment at the review-level \cite{pontiki2016semeval}. As such, document-level ABSA considers the general concepts that summarize the sentiment in the text, while sentence-level ABSA considers each mention of an aspect individually. The task of ABSA can be further split up into three tasks: aspect detection/extraction, sentiment classification, and sentiment aggregation \cite{schouten2015survey}. Aspect extraction involves identifying the aspects present in a text, the classification step consists of assigning a sentiment label or score to the extracted aspects, and the aggregation step amounts to summarizing the aspect sentiment classifications. In this survey, we focus on the sentiment classification step, which is generally referred to as aspect-based sentiment classification (ABSC).

Various surveys on ABSA have already been published \cite{schouten2015survey, pavlopoulos2014aspect}. However, it seems that a survey focused specifically on ABSC would be more effective at providing an in-depth discussion and evaluation of ABSC models. The only survey dedicated solely to ABSC is \cite{zhou2019deep}, which provides an overview of deep learning techniques. While deep learning models are currently state-of-the-art for ABSC, we would argue that a broader scope would allow for a more effective evaluation of the current and future state of research in ABSC. For example, a significant part of ABSC is the research on approaches that incorporate knowledge bases into classification models. Furthermore, various important deep learning models, such as transformer-based models \cite{vaswani2017attention}, are missing from previous surveys. For these reasons, this survey presents a comprehensive overview of the state-of-the-art ABSC models. For this purpose, we propose a novel taxonomy of ABSC models, which is currently missing from the literature. The taxonomy categorizes ABSC models into three major categories: knowledge-based methods, machine learning models, and hybrid models. Based on the structure of this taxonomy, we discuss and compare the architectures of the various ABSC models using both technical and intuitive explanations. Furthermore, we also provide summarizing overviews of the performances of various ABSC models on a larger scale than previous surveys. Last, we identify trends in the research on ABSC and use these findings to discuss ways in which the field of ABSC can be advanced in the future.

The sections of this survey are based on the main steps taken when designing ABSC methods. In Section \ref{sec:Inputs}, we discuss the different ways of representing the inputs for ABSC. In Section \ref{sec:Evaluation}, techniques for evaluating the model performance are reviewed. In Section \ref{sec:Methods}, various ABSC models are presented according to a proposed taxonomy. We discuss how these models use the preprocessed inputs to produce the desired classification output, and compare the performances of various models found in the literature. In Section \ref{sec:Related}, additional topics related to ABSC are discussed. In Section \ref{sec:Conclusion}, we give our conclusions and discuss directions for future research.

\section{Input Representation}\label{sec:Inputs}

In this section, we explain in detail the input representations necessary for ABSC. We start by introducing some definitions in accordance with Schouten and Frasincar \cite{schouten2015survey}. Given a corpus $C$ containing the records $R_1, \dots, R_{n_R}$, ABSA can be formally defined as finding all quadruples $(y, a, h, t)$ \cite{liu2012sentiment} for each record $R_j$, where $y$ is the sentiment, $a$ represents the target aspect for the sentiment, $h$ is the holder, or the individual expressing the sentiment, and $t$ represents the time that the sentiment was expressed \cite{schouten2015survey}. A record is defined as an individual piece of text in the corpus, which can be a single phrase, a sentence, or a large body of text, like a document. Furthermore, in general, most methods are concerned with finding $(y, a)$, the aspects and the corresponding sentiments. Since this survey is only focused on the classification step of ABSA, we assume the target aspects $a$ are already identified in the text. As such, ABSC models are focused only on finding the sentiments $y$ corresponding to the given aspects $a$. For example, consider the following record that takes the form of a restaurant review: \textit{``The atmosphere was fantastic, but the food was bland"}. This sentence contains two aspects \textit{``atmosphere"} and \textit{``food"}, which we assume have been identified in a prior aspect extraction phase. The eventual goal is to use an ABSC model to determine a sentiment classification for each of these aspects. The correct sentiments, in this case, would be \textit{``positive"} and \textit{``negative"}, respectively. However, before one can attempt this task, input representations need to be constructed.

Since a piece of text can generally not directly be used as input for a classification model, the aspect and corresponding context must be represented through numeric features. Note that, in aspect-based, or feature-based, sentiment analysis, the term `features' is sometimes used to describe the aspects of, for example, a product. However, `features' in this context refers not to the aspects themselves, but to the features of the data. Before an ABSC model can be implemented, a preprocessing phase is necessary to construct a numeric representation \cite{haddi2013preprocess, goldberg2017neural}. The features that are used as inputs are a crucial part of the classification process since they determine what information the ABSC model has access to. It is important to understand how to represent a piece of a text in such a way that the classification model has access to as much information as possible without including redundant or irrelevant features such that it can perform optimally. Input representations for ABSC generally consist of three characteristics: \textbf{context}, \textbf{dimensionality}, and \textbf{feature types}. We discuss these in Subsection \ref{sec:InputContext}, Subsection \ref{sec:InputDimensionality}, and Subsection \ref{sec:InputTypes}, respectively.

\subsection{Context}\label{sec:InputContext}
Given a record $R_j$, we define the \textbf{context} as the subset of words that is considered as the input. If the text only contains a single aspect, then one can choose to simply consider all words. However, if the record contains multiple aspects, a representation needs to be developed for each aspect individually by, for example, taking a subset of the words for each aspect. The methods of representing the input for each aspect depend on whether a target phrase is present in the text. As such, we make a distinction in the input representation techniques for \textbf{explicit} and \textbf{implicit} aspects.

Explicit aspects are generally the most common type of aspects. An example of such an aspect can be seen in the review sentence \textit{``The price of this phone is very high."}, where the aspect is explicitly stated to be \textit{``price"}. The simplest method of determining the context for such an explicit aspect is by employing a window around the target phrase and developing an input representation based only on the words in the window. For example, Guha et al. \cite{guha2015siel} only consider the aspect itself, the three words to the left of the aspect, and the three words to the right of the aspect. However, such methods based on the physical proximity may not be optimal since the words expressing the sentiment may be far removed from the aspect. As such, more robust methods do not rely on the physical distance between words. For example, one can use the grammatical dependencies to determine which words in the record are related to the aspect and should therefore be considered in the context \cite{thet2010aspect}. Another technique for determining the context is the use of text kernels to express the distances based on the relations between words. Nguyen and Shirai \cite{nguyen2015treekernel} implement tree kernels for relation extraction to determine which words they consider in the analysis.

An example of a sentence with an implicit aspect is \textit{``This phone is really expensive."}, where the aspect is again \textit{``price"}, but the aspect is not directly mentioned in the text. Implicit aspects must be treated differently since we do not have a target phrase to center a window around or to determine distances to words with. If we assume the record contains only one aspect, such as in the given example sentence, then this is not a major problem as one can simply consider the whole sentence for the input representation. However, as Dosoula et al. \cite{dosoula2016sentiment} argue, reviews may often contain multiple implicit aspects, even within the same sentence. For instance, the previously given example can be extended with another implicit aspect: \textit{``This phone is really expensive, but also very fast"}. Dosoula et al. \cite{dosoula2016sentiment} present different methods based on aspect proxy words to determine the context for each aspect in a sentence.

\subsection{Dimensionality}\label{sec:InputDimensionality}
The \textbf{dimensionality} of the input representation is directly determined by the type of model used in the subsequent sentiment classification step. Some models, such as support vector machines and decision trees, work only with a single vector representation, whereas other models, such as recurrent neural networks, can process a set of vectors or a data matrix. Suppose record $R_j$ contains a single aspect and we wish to represent the entire record using numeric features. Then, depending on the ABSC models used, record $R_j$ can be represented using a single vector $\bm x_j \in \mathbb{R}^{d_x}$, or a matrix $\bm X_j \in \mathbb{R}^{d_x \times n_j}$, where $d_x$ represents the number of features used for the representation, and $n_j$ represents the number of words in record $R_j$. A single vector representation of the record can be interpreted as a record embedding where each element of the vector indicates the presence of a feature. For example, whether certain sentiment-indicating words or phrases are present. On the other hand, a data matrix can provide more detailed information since it does not have to summarize all information into a single vector. Each column of the matrix is an embedding containing features representing a part of the record, such as a sentence, word, or character. The dimensionality of the inputs generally influences the types of features used, as some feature types work better with a vector representation, whereas others can be used more effectively with a matrix representation. The next subsection discusses this in further detail.

\subsection{Feature Types}\label{sec:InputTypes}
When representing text as a single vector, classical text classification \textbf{feature types} can be used. Arguably one of the simplest and most popular ways of representing a sentence or piece of text is via a bag-of-words (BoW) representation. A BoW vector representation is a feature vector where each element represents a word. It is called a ``bag" because we simply congregate the words together and disregard the structure of the text. Again, suppose we wish to represent record $R_j$ using a single vector $\bm x_j$. We start by building a vocabulary that contains every unique word used in any of the records in our corpus $C$. For a BoW representation, each element $x_{i,j}$ in vector $\bm x_j$ represents a unique word from the vocabulary. The simplest BoW vector is a binary vector where each element $x_{i,j}$ in vector $\bm x_j$ is set equal to 1 if the corresponding word occurs in $R_j$, and 0 otherwise. A more popular version is to also consider the frequency of the words by setting each element $x_{i,j}$ equal to the number of times the corresponding word occurs in $R_j$ \cite{salton1968tf}. However, some words are simply used more often than others, which can cause a bias towards certain words in the word counts. Thus, instead of using the direct word counts, one can also implement the term frequency-inverse document frequency (TF-IDF) measure \cite{jones1972statistical}. This measure is essentially the word frequency in a record scaled by how often the word appears in all records. For word $w_i$ in the vocabulary, the TF-IDF can be calculated using Equation (\ref{eq:TF-IDF_BoW}).
\begin{equation}\label{eq:TF-IDF_BoW}
    \def\sss{\scriptscriptstyle}
    \setstackgap{L}{8pt}
    \def\stacktype{L}
    x_{i,j} = \frac{n_{i,j}}{n_j} \times \text{log}\frac{n_R}{|\{R_j \in C: w_i \in R_j \}|},
\end{equation}
where $n_{i,j}$ denotes the number of times word $w_i$ occurs in record $R_j$, $n_j$ represents the number of words in record $R_j$, $n_R$ denotes the total number of records, and $C$ indicates the set of records. As previously mentioned, for ABSC, one should preferably not use a singular text representation when multiple aspects are present in the record. As such, one can simply create a BoW representation based only on the words considered to be in the context.

\begin{table}
\caption{Overview of various types of features.}
\label{Table:FeatureTypes}
\begin{center}
\resizebox{230pt}{!}{%
\begin{tabular}{p{100pt}|p{150pt}}
\textbf{Feature Types} & \textbf{Feature Examples} \\ \hline
Word Features & Bag-of-words \cite{scott1998text}, word embeddings \cite{mikolov2013efficient}, $n$-grams \cite{pak2010twitter}\\ \hline
Syntactic Features & Part-of-speech tagging \cite{ratnaparkhi1996maximum, dragoni2018combining} and dependency parsing \cite{dragoni2018combining, kubler2009dependency, mukherjee2012feature, schouten2015benefit} \\ \hline
Proximity-Based Features & Proximity relative to sentiment words \cite{mullen2004sentiment, hasan2011proximity} or target words \cite{jebbara2016aspect} \\ \hline
Semantic Features & Sentiment scores of context words \cite{hatzivassiloglou1997predicting, turney2002SO}\\ \hline
Morphological Features & Lexemes and lemmas \cite{alsmadi2019enhancing, schouten2015benefit, abdul2014samar} \\
\end{tabular}
}
\end{center}
\end{table}

BoW features work well for ABSC but may come with certain problems that make the classification process difficult. Firstly, a large corpus can contain a substantial number of different words, meaning that the vocabulary and corresponding feature vectors will be very large if one would represent every single word with an element. Thus, methods have been developed to select the most important words that can be used for features. For example, based on a list or dictionary, one can filter out the words that generally hold no meaningful or semantic value, such as stop words. Even though such techniques can significantly reduce the number of features, the vectors generally remain large. Because of this, an important characteristic of ABSC models is that they should typically be able to handle high-dimensional vectors well, which is discussed in more detail in Section \ref{sec:Methods}. Secondly, as previously mentioned, the structure of the words is completely disregarded, making it difficult to capture relations between the aspect and its context. One solution is to include $n$-grams, but this further exacerbates the previously mentioned high dimensionality problem. Another solution is to include features that define the relations between the words. Additional feature types are generally used and can take a variety of forms to include different types of information. For example, Al-Smadi et al. \cite{alsmadi2019enhancing} enhance a TF-IDF BoW representation with additional morphological, syntactic, and semantic features. Similarly, Mullen and Collier \cite{mullen2004sentiment} implement semantic, syntactic, and proximity-based features. Table \ref{Table:FeatureTypes} provides an overview of a variety of feature types with the corresponding examples. 
 
Suppose we now wish to represent the record $R_j$ using a matrix $\bm X_j$. In this case, the idea is that each word in the text is represented by a vector, which is stored as a column in $\bm X_j$. A standard choice, in this case, is the use of word embeddings. For an overview of the many types of word embeddings, we refer to \cite{camacho2018word}. Some examples are: \textit{GloVe} \cite{pennington2014glove}, \textit{fastText} \cite{grave2018learning}, and \textit{Word2Vec} \cite{mikolov2017advances}. Each of these embedding types attempts to represent a word's meaning using a vector of limited size. While pre-trained word embeddings are often used, it is also possible to train word embeddings along with the ABSC model. Additionally, the previously mentioned embedding models are all non-contextual, which means that an embedding is produced for each word independently from the other words. Yet, it is known that a word can have multiple meanings depending on the context that it is used in. Therefore, another solution is to use contextual word embeddings, such as embeddings based on \textit{ELMo} \cite{peters2018deep} or \textit{BERT} \cite{devlin2018bert}. These word embedding models produce different vectors for words depending on the surrounding context. For example, the word ``playing" has a different meaning in ``playing tennis" than in ``playing piano" and will therefore get a different word embedding assigned to it. Furthermore, there are also word embeddings made specifically for sentiment analysis \cite{7296633}. Word embeddings are a powerful representation tool on their own, but one can still add additional feature types like the examples provided in Table \ref{Table:FeatureTypes}. For example, in \cite{alsmadi2018rnnvsvm}, word embeddings are enhanced with syntactic and semantic features. The input representations discussed in this section are the basis for the classification models discussed in Section \ref{sec:Methods}. In the rest of this paper, we assume that input representations have already been constructed beforehand. We only further elaborate on the input representation if the classification model modifies the input representations. Furthermore, in the rest of this paper, for clarity purposes, we do not use record-specific or aspect-specific subscripts. 

\section{Performance Evaluation}\label{sec:Evaluation}
The goal of an ABSC model is to produce an output using the input representations. Thus, given a record $R$ containing an aspect $a$, an ABSC model produces a label $\hat{y} \in \mathbb{R}^{1}$ using the feature vector $\bm x \in \mathbb{R}^{d_x}$ or matrix $\bm X \in \mathbb{R}^{d_x \times n_x}$, where $d_x$ represents the number of features used, and $n_x$ indicates the number of words considered to be in the context. The effectiveness of ABSC models can be compared by evaluating the output $\hat{y}$ using a variety of performance measures. Each of these measures highlights certain strengths and weaknesses of the classification models. In this section, we represent various techniques for evaluating ABSC models. These measures are used to compare and contrast the various ABSC approaches in Section \ref{sec:Methods}. The most commonly used performance measures for ABSC and ABSA, in general, are the well-known accuracy, precision, recall, and $F_1$-measure \cite{schouten2015survey}. These measures evaluate the sentiment classification $\hat{y}$ by comparing it to the true sentiment label $y$ of the aspect. Models that achieve high values for these performance measures, will produce higher-quality predictions. The accuracy measures the ratio of correctly classified aspects to the total number of aspects in the dataset. This is intuitively clear, as it is a direct indication of how well the model predicts. In a binary classification setting, a simple baseline to use is a random classifier. A fully random classifier would obtain an expected accuracy of 0.5, which any predictive model should be able to outperform. Yet, the accuracy measure is often not a valid indicator of performance in imbalanced datasets. If a dataset contains two classes, where 90\% of the records are of one class, then always predicting that class will provide a high accuracy of 0.9. As such, other measures like the precision and recall are used that provide more useful evaluations of classification performance in imbalanced datasets. One can aggregate the precision and recall for each class to obtain a measure of the model performance. This can be done by evaluating the mean of the measures (macro-averaging), or by aggregating based on the contributions of all classes (micro-averaging). A similar process can be used when aggregating the results from multiple different datasets. One can either take the mean of the performance measures from the various datasets (macro-averaging), or one can aggregate based on the contributions of the datasets (micro-averaging). As the precision and recall focus on different parts of the model performance, the F$_1$-measure is typically used to summarize the information.

The accuracy, precision, recall, and F$_1$-measure are used in most of the works discussed in this survey. However, there are various alternative measures that are occasionally used as well. For example, the mean squared error (MSE) and the ranking loss. Suppose we have a training dataset consisting of the $N$ feature vectors $\bm x_1, \dots, \bm x_N$ with the corresponding labels $y_1, \dots, y_N$. Each of the feature vectors $\bm x_1, \dots, \bm x_N$ represents an aspect with the corresponding context in a record, as explained in Section \ref{sec:Inputs}. The corresponding labels $y_1, \dots, y_N$ indicate the true sentiment expressed towards the aspects. An ABSC model is used to produce the label predictions $\hat{y}_1, \dots, \hat{y}_N$. Since the labels for sentiment analysis are generally considered to be ordinal, the MSE can be calculated as follows:
\begin{equation}\label{equation:MSE}
    \def\sss{\scriptscriptstyle}
    \setstackgap{L}{8pt}
    \def\stacktype{L}
    \text{MSE} = \frac{1}{N}\sum_{i=1}^{N}(\stackunder{\hat{y}_i}{\sss 1 \times 1} - \stackunder{y_i}{\sss 1 \times 1})^2.
\end{equation}
Due to the square, the MSE penalizes large errors more than small errors. An alternative is the ranking loss \cite{crammer2002pranking} that punishes small and large errors in a more equal manner. The ranking loss measure is closely related to the mean absolute error and can be calculated according to Equation (\ref{equation:MAE}).
\begin{equation}\label{equation:MAE}
    \def\sss{\scriptscriptstyle}
    \setstackgap{L}{8pt}
    \def\stacktype{L}
    \text{Ranking Loss} = \frac{1}{N}\sum_{i=1}^{N}|\stackunder{\hat{y}_i}{\sss 1 \times 1} - \stackunder{y_i}{\sss 1 \times 1}|,
\end{equation}
Both the Ranking Loss and the MSE measures are used to measure the errors in the classification predictions. As such, the lower the values attained for these measures, the better the model performs. A final performance measure is the area under the Receiver Operating Characteristics (ROC) curve, abbreviated as AUC. The ROC curve plots the recall against the true positive rate. The area under the curve (AUC) is then a measure of how well the model can separate classes, meaning that the higher the AUC value, the better the performance of the model.

\section{Sentiment Classification}\label{sec:Methods}

\begin{figure}
    \centering
    \includegraphics[scale=0.7]{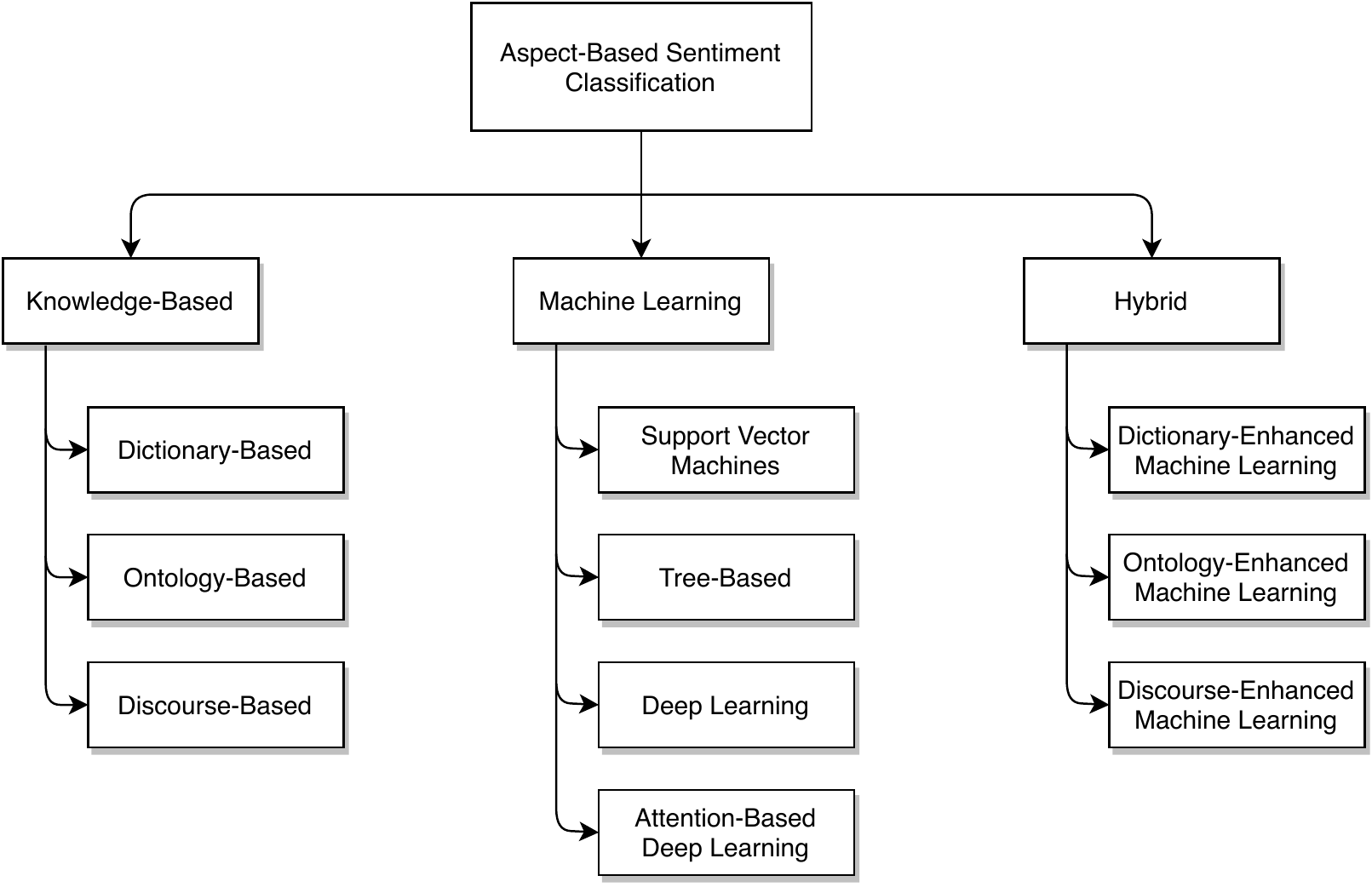}
    \caption{Taxonomy of ABSC methods.}
    \label{fig:Taxonomy}
\end{figure}

ABSC models can generally be classified into three major categories: knowledge-based approaches, machine learning models, and hybrid models. In Figure \ref{fig:Taxonomy}, a taxonomy consisting of these categories and their sub-categories is presented. The taxonomy categories are explained in more detail in Subsection \ref{sec:MethodsKnowledgeBased}, Subsection \ref{sec:MethodsMachineLearning}, and Subsection \ref{sec:MethodsHybrid}, respectively. Each of these subsections contains an overview of various prominent ABSC models for that model type in the form of a summarizing table. Each table consists of columns detailing the model, the types of data used, and the various performance measures reported for every model. A row corresponding to a model can consist of multiple entries if datasets from different domains are used. Additionally, we report results from other works that implement the same model using different datasets. When a model is reimplemented in another paper, we indicate this beneath the entry. Unavailable information is indicated by a ``-''. Since we are unable to include all results reported by each paper, we take several steps to summarize the results. First of all, we only include the results of the best model architecture presented in each paper. Secondly, when multiple datasets are used in the same domain, we average the results across those datasets. The only exception is when a model is reimplemented in another paper since we cannot guarantee that the model implementation is the same. We have included the references for all datasets in the tables for easier model comparisons. General characteristics of some ABSC datasets are provided in Table \ref{Table:Data}. Note that this table only includes the most used domains and languages from the listed datasets. For example, the SemEval-2016 dataset contains many additional languages, such as Dutch, Chinese, and Turkish.

\begin{table*}
\caption{Overview of datasets used for ABSC.}
\label{Table:Data}
\begin{center}
\resizebox{430pt}{!}{%
\begin{tabular}{p{105pt}|p{120pt}|p{60pt}|p{40pt}|p{40pt}|p{40pt}|p{40pt}}
\textbf{Dataset} & \textbf{Domain(s)} & \textbf{Language(s)} & \textbf{Positives} & \textbf{Neutrals} & \textbf{Negatives} & \textbf{Total} \\ \hline
SemEval-2014 \cite{pontiki-etal-2014-semeval}  & Reviews (Electronics) \newline Reviews (Restaurants) & English \newline English & 1328 \newline 2894 & 629 \newline 829 & 994 \newline 1001 & 2951 \newline 4724 \\ \hline
SemEval-2015 \cite{pontiki2015semeval}  & Reviews (Electronics) \newline Reviews (Restaurants) \newline Reviews (Hotels) & English \newline English \newline English & 1644 \newline 1652 \newline 243 & 185 \newline 98 \newline 12 & 1094 \newline 749 \newline 84 & 2923 \newline 2499 \newline 339 \\ \hline
SemEval-2016 \cite{pontiki2016semeval}  & Reviews (Electronics) \newline Reviews (Restaurants) \newline Reviews (Hotels) & English \newline English \newline Arabic & 1540 \newline 1802\newline 7705 & 154 \newline 104\newline 852 & 869 \newline 623\newline 4556 & 2563 \newline 2529 \newline 13113 \\ \hline
Twitter Data \cite{dong2014adaptive}  & Social Media (Various) & English & 1735 & 3470  & 1735 & 6940 \\ \hline
SentiHood \cite{saeidi-etal-2016-sentihood}  & Social Media (Neighborhoods) & English & 2460 & 0  & 1216 & 3676 \\ \hline
Hindi Reviews \cite{akhtar2016aspect_h}  & Reviews (Various) & Hindi & 1986 & 1914  & 569 & 4469 \\ \hline
Indonesian Marketplace \cite{8285850} & Reviews (Various) & Indonesian & 7982 & 0  & 1677 & 9659 \\ \hline
FiQA-2018 \cite{10.1145/3184558.3192301} & Social Media (Finance) & English & 722 & 13  & 378 & 1113 \\ \hline
\end{tabular}
}
\end{center}
\end{table*}

\begin{table*}
\caption{Overview of prominent knowledge-based ABSC models and their reported performances. If no performance measures are displayed, then no quantitative analysis was provided for ABSC in the original paper.}
\label{Table:KB}
\begin{center}
\resizebox{430pt}{!}{%
\begin{tabular}{p{115pt}|p{65pt}|p{118pt}|p{40pt}|p{40pt}|p{28pt}|p{18pt}|p{50pt}}
\textbf{Reference} & \textbf{Category} & \textbf{Domain(s)} & \textbf{Accuracy} & \textbf{Precision} & \textbf{Recall} & \textbf{F$_1$} & \textbf{Alternative \newline Measures} \\ \hline
Hu and Liu (2004) \cite{hu2004mining} & Dictionary-Based & Reviews (Electronics) \cite{amazon, cnet} & 0.842 & 0.642 & 0.693 & 0.667 & - \\ \hline
Zhu et al. (2009) \cite{zhu2009multiaspect} & Dictionary-Based & Reviews (Restaurants) \cite{dianping} & - & 0.755 & 0.755 & 0.755 & - \\ \hline
Moghaddam and Ester (2010) \cite{moghaddam2010opinion} & Dictionary-Based & Reviews (Electronics) \cite{epinions} & - & - & - & - & Ranking Loss: \newline 0.49 \\ \hline
Eirinaki et al. (2012) \cite{eirinaki2012feature} & Dictionary-Based & Reviews (Various) \cite{sentdata} & - & - & - & - & - \\ \hline
Zhou and Chaovalit (2008) \cite{zhou2008ontology} & Ontology-Based & Reviews (Movies) \cite{imdb} & 0.722 & - & - & - & - \\ \hline
Kontopoulos et al.\newline (2013) \cite{kontopoulos2013ontology} & Ontology-Based & Social Media (Electronics) \cite{twitter} & - & - & - & - & -  \\ \hline
Nie et al. (2013) \cite{nie2013opinion} & Ontology-Based & Reviews (Electronics) \cite{jd} & - & 0.587 & 0.625 & 0.605 & - \\ \hline
Hoogervorst et al. (2016) \cite{hoogervorst2016aspect} & Discourse-Based & Reviews (Electronics) \cite{pontiki2015semeval}\newline Reviews (Restaurants) \cite{pontiki-etal-2014-semeval, pontiki2015semeval} & -\newline - & 0.670 \newline 0.670  & 0.670 \newline 0.670 & 0.670 \newline 0.670 & -\newline - \\ \hline
Sanglerdsinlapachai et al. (2016) \cite{sanglerdsinlapachai2016exploring} & Discourse-Based & Reviews (Electronics) \cite{cnet} & 0.633 & - & - & - & - \\ \hline
Dragoni et al. (2018) \cite{dragoni2018combining} & Discourse-Based & Reviews (Electronics) \cite{pontiki2015semeval}\newline Reviews (Restaurants) \cite{pontiki2015semeval} & 0.859\newline 0.779 & - \newline -  & - \newline - & - \newline - & -\newline - \\
\end{tabular}
}
\end{center}
\end{table*}

\subsection{Knowledge-Based}\label{sec:MethodsKnowledgeBased}
Knowledge-based methods, also known as symbolic AI, are approaches that make use of a knowledge base. Knowledge bases are generally defined as stores of information with underlying sets of rules, relations, and assumptions that a computer system can draw upon. Knowledge-based methods are heavily related to the input representation since these approaches often use a knowledge base to define features. One of the advantages of knowledge-based methods is interpretability. Namely, it is generally easy to identify the information used to produce the model output. The underlying mechanisms of knowledge-based methods are typically relatively simple, which allows for highly transparent approaches to ABSC. Knowledge-based methods require no training time, but the construction of knowledge bases can take considerable time. We discuss the following three knowledge-based approaches: dictionary-based, ontology-based, and discourse-based methods. The performances of various knowledge-based methods are displayed in Table \ref{Table:KB}.

\subsubsection{Dictionary-Based}\label{sec:MethodsDictionaryBased}
Early methods for ABSC were mostly based on dictionaries. Given a record $R$ containing an aspect $a$, dictionary-based methods construct a feature vector $\bm x$ using a dictionary, where each element $x_i$ represents a sentiment score, or orientation, of a word in the context regarding the aspect. Various dictionaries can be used for ABSC, such as \textit{WordNet} \cite{miller1995wordnet} and \textit{SentiWordNet} \cite{baccianella2010sentiwordnet}. These dictionaries define sets of words and the linguistic relations between them. For example, \textit{WordNet} \cite{miller1995wordnet} groups nouns, verbs, adjectives, and adverbs into so-called \textit{synsets}: sets of synonyms, or groups of words that have the same meaning. One can exploit these relations for ABSC since words generally portray the same sentiment as their synonyms \cite{hu2004mining}. As such, we start with a set of \textit{seed words} for which the sentiment is known. For example, one can use \textit{``good"}, \textit{``fantastic"}, and \textit{``perfect"} as positive seed words, and \textit{``bad"}, \textit{``boring"}, and \textit{``ugly"} as negative seed words. These seed words can then be used to determine the sentiment of words surrounding an aspect via the synsets defined in a dictionary, like \textit{WordNet} \cite{miller1995wordnet}. Words that are synonyms of a positive seed word, or antonyms of a negative seed word, will receive a positive sentiment score. \textit{SentiWordNet} \cite{baccianella2010sentiwordnet} is in part based on this idea. This dictionary generates a sentiment label (\textit{``positive"}, \textit{``neutral"}, or \textit{``negative"}) for each of the synsets using a combination of a seed set expansion approach and a variety of classification models.

After determining sentiment scores for the context words, a method needs to be implemented to determine a sentiment output. For example, the authors of \cite{hu2004mining} examine whether there are mostly negative or positive words present in the context. In \cite{hu2004mining}, the sentiment scores for the context words are encoded as 1 for positive, -1 for negative, and 0 for neutral. Then, the sentiment classification can be determined by summing the elements of $\bm x$. If the sum is positive, a positive label is returned, and a negative label otherwise. Similarly, in \cite{eirinaki2012feature}, the context words are assigned a sentiment score in the range $[-4, 4]$. The sentiment polarity of the aspect is then determined based on the average score per opinion word relating to the aspect.

\subsubsection{Ontology-Based}\label{sec:MethodsOntologyBased}
An ontology is generally defined as an \textit{``explicit, machine-readable specification of a shared conceptualization"} \cite{gruber1995toward, studer1998knowledge}, and defines a set of entities and relations corresponding to their properties. The main difference between an ontology and a dictionary is that a dictionary captures linguistic relations between words, while an ontology represents relations between real entities. These relations can be used for determining which words in the record are important for determining the sentiment towards the aspect. For ABSC, one can either use an existing ontology or create one based on the domain at hand \cite{kontopoulos2013ontology}. Examples of existing ontologies are the semantically interlinked web communities (SIOC) ontology \cite{SIOC}, which is an ontology that captures data from online community websites, and the ontology of emotions proposed in \cite{roberts2012empatweet}. Yet, rather than relying on existing ones, most researchers choose to create their own ontology. This is because finding an ontology that relates well to a specific domain is difficult since many existing ontologies typically capture rather generic concepts. Designing ontologies is often done using specific methods, such as formal concept analysis \cite{obitko2004ontology} or the \textit{OntoClean} methodology \cite{guarino2002evaluating}. Ontologies are often manually created to ensure the accuracy of the captured relations \cite{mevskele2020aldonar}, but this process is rather time-consuming. As such, methods for creating ontologies semi-automatically or even completely automatically can be vital when creating larger ontologies. For example, in \cite{10.1007/978-3-319-12024-9_4}, an ontology for ABSC is created semi-automatically using a proposed semantic asset management workbench (SAMW). A method for fully automatically creating ontologies is proposed in \cite{1179189}.

Ontologies capture the structure of objects in a domain. These relations can be used to determine the context corresponding to the aspect \cite{kontopoulos2013ontology}. Then, a method can be used to determine the sentiment label using the context obtained using the ontology. For example, by using a dictionary-based sentiment classifier \cite{zhou2008ontology}. While regular ontologies are useful tools for defining relations between objects, to further facilitate sentiment analysis, sentiment information can be incorporated into the ontology. Such sentiment ontologies specifically define sentiment relations between words or entities. For example, in \cite{nie2013opinion}, a sentiment ontology tree is constructed that links product aspects to opinion words with the corresponding sentiment scores. In \cite{wojcik2014ontology}, a sentiment ontology is constructed where the sentiment relations are defined using the \textit{SentiWordNet} dictionary \cite{baccianella2010sentiwordnet}. Zhuang et al. \cite{zhuang2019soba} propose a semi-automated ontology builder for ABSA that constructs an ontology based on sentiment relations. When classifying an aspect, one can link the words in the context to the concepts and relations in the sentiment ontology, and summarize the sentiment relations to produce a sentiment classification. The semi-automatic method proposed in \cite{zhuang2019soba} mainly focuses on word frequencies, due to the limited amount of domain data. In \cite{dera2020sasobus}, another semi-automatic ontology builder is proposed that is based on synsets from the \textit{WordNet} \cite{miller1995wordnet} dictionary. Lastly, ten Haaf et al. \cite{tenhaaf2021websoba} propose a semi-automatic ontology builder that is based on word embeddings produced using the \textit{word2vec} \cite{mikolov2017advances} method.

\subsubsection{Discourse-Based}\label{sec:MethodsDiscourseBased}
Another type of knowledge base that can be used for ABSC is a discourse tree based on rhetorical structure theory (RST) \cite{mann1988rhetorical}. RST can be used to define a hierarchical discourse structure within a record to categorize phrases into elementary discourse units (EDU). Similar to ontologies, discourse trees define sets of relations that can be used to determine which words are important when assigning a sentiment classification to the aspect. For a record $R$, a discourse tree is constructed that defines the hierarchical discourse relations in the record. To determine the context concerning an aspect $a$ in the record, the authors of \cite{hoogervorst2016aspect} use an RST-based method that produces a sub-tree of the discourse tree that contains the discourse relations specifically regarding the aspect. This sub-tree is named the context tree and can be used to determine the sentiment classification of the aspect. Since the context tree does not contain any sentiment information, the authors of \cite{hoogervorst2016aspect} use a dictionary-based method to label the leaf nodes of the context tree with sentiment orientation scores for sentence-level ABSC. To determine the sentiment classification of the aspect, one can simply evaluate the sum of the sentiment scores defined for the context tree. In \cite{sanglerdsinlapachai2016exploring}, a similar technique is used with a more sophisticated method of aggregating the scores.

There are other types of discourse structure theories besides RST \cite{HOU2020113421}. One example of this is cross-document structure theory (CST) \cite{10.3115/1117736.1117745}, which can be used to analyse the structure and discourse relations between groups of documents. Cross-document structure analysis can be useful in social media analysis since social media posts are short documents that are typically highly interrelated. An interesting example of this is the SMACk system \cite{dragoni2016smack}, which analyses a cross-document structure based on abstract argumentation theory \cite{DUNG1995321}. For example, products often receive multiple reviews from various users that may respond to each other. The SMACk system analyses the relations between the different arguments presented in the reviews to improve the aggregation of the sentiment expressed towards the different aspects \cite{dragoni2018combining}.

\subsection{Machine Learning}\label{sec:MethodsMachineLearning}

As opposed to knowledge-based methods that exploit knowledge bases to achieve a sentiment classification, machine learning models, also known as subsymbolic AI, use a training dataset of feature vectors and corresponding correct labels. The machine learning model is trained to extract patterns from the data that can be used to distinguish between sentiment classes. There are many types of machine learning models, which can be grouped into the following categories: support vector machines, tree-based models, deep learning models, and attention-based deep learning models. The performances of various machine learning models are presented in Table \ref{Table:ML}.

\begin{table*}
\caption{Overview of prominent machine learning ABSC models and their reported performances. DT = ``Decision Tree" and RF = ``Random Forest".}

\label{Table:ML}
\begin{center}
\resizebox{430pt}{!}{%
\begin{tabular}{p{110pt}|p{100pt}|p{120pt}|p{40pt}|p{40pt}|p{28pt}|p{20pt}|p{45pt}}
\textbf{Reference} & \textbf{Category} & \textbf{Domain(s)} & \textbf{Accuracy} & \textbf{Precision} & \textbf{Recall} & \textbf{F$_1$} & \textbf{Alternative \newline Measures} \\ \hline
Jiang et al. (2011) \cite{jiang2011target} & SVM & Social Media (Various) \cite{twitter}\newline Social Media (Various) \cite{dong2014adaptive} \newline (reimplemented in \cite{vo2015target}) & 0.682\newline 0.634 & -\newline - & -\newline - & -\newline 0.633 & -\newline - \\ \hline
Yu et al. (2011) \cite{yu2011aspect} & SVM & Reviews (Electronics) \cite{cnet, viewpoints, reevoo, gsmarena} & - & - & - & 0.787 & - \\ \hline
Pannala et al. (2016) \cite{pannala2016supervised} & SVM & Reviews (Electronics) \cite{pontiki2015semeval}\newline Reviews (Restaurants) \cite{pontiki2015semeval} & 0.732\newline 0.823 & -\newline - & -\newline - & -\newline - & -\newline - \\ \hline
Akhtar et al. (2016) \cite{akhtar2016aspect} & SVM & Reviews (Electronics) \cite{akhtar2016aspect}\newline Reviews (Mobile Apps) \cite{akhtar2016aspect}\newline Reviews (Holidays) \cite{akhtar2016aspect}\newline Reviews (Movies) \cite{akhtar2016aspect} & 0.511\newline 0.421\newline 0.606\newline 0.916 & -\newline -\newline -\newline - & -\newline -\newline -\newline - & -\newline -\newline -\newline - & -\newline -\newline -\newline - \\ \hline
Akhtar et al. (2016) \cite{akhtar2016aspect_h} & SVM & Reviews (Various) \cite{akhtar2016aspect_h} & 0.541 & - & - & - & - \\ \hline
Hegde and Seema (2017) \cite{7972395} & SVM & Reviews (Various) \cite{7972395} & 0.758 & 0.762 & 0.755 & 0.758 & - \\ \hline
De Fran\c{c}a Costa and da Silva (2018) \cite{10.1145/3184558.3191828} & SVM & Social Media (Finance) \cite{10.1145/3184558.3192301} & - & - & - & - & MSE = 0.151 \\ \hline
Al-Smadi et al. (2019) \cite{alsmadi2019enhancing} & SVM & Reviews (Hotels) \cite{pontiki2016semeval} & 0.954 & - & - & - & AUC = 0.989 \\ \hline
Akhtar et al, (2016) \cite{akhtar2016aspect} & Tree-Based (DT) & Reviews (Electronics) \cite{akhtar2016aspect}\newline Reviews (Mobile Apps) \cite{akhtar2016aspect}\newline Reviews (Holidays) \cite{akhtar2016aspect}\newline Reviews (Movies) \cite{akhtar2016aspect} & 0.545\newline 0.480\newline 0.652\newline 0.916 & -\newline -\newline -\newline - & -\newline -\newline -\newline - & -\newline -\newline -\newline - & -\newline -\newline -\newline - \\ \hline
Hegde and Seema (2017) \cite{7972395} & Tree-Based (DT) & Reviews (Various) \cite{7972395} & 0.806 & 0.729 & 0.792 & 0.759 & - \\ \hline
Al-Smadi et al. (2019) \cite{alsmadi2019enhancing} & Tree-Based (DT) & Reviews (Hotels) \cite{pontiki2016semeval} & 0.949 & - & - & - & AUC = 0.996 \\ \hline
Gupta et al. (2014) \cite{gupta-ekbal-2014-iitp} & Tree-Based (RF) & Reviews (Electronics) \cite{pontiki-etal-2014-semeval}\newline Reviews (Restaurants) \cite{pontiki-etal-2014-semeval} & 0.671\newline 0.674 & -\newline - & -\newline - & -\newline - & -\newline - \\ \hline
Tang et al. (2016) \cite{tang-etal-2016-effective} & Deep Learning (RNN) & Social Media (Various) \cite{dong2014adaptive} & 0.715 & - & - & 0.695 & - \\ \hline
Ruder et al. (2016) \cite{ruder2016hierarchical} & Deep Learning (RNN) & Reviews (Electronics) \cite{pontiki2016semeval}\newline Reviews (Restaurants) \cite{pontiki2016semeval} \newline Reviews (Hotels) \cite{pontiki2016semeval}& 0.790 \newline 0.807 \newline 0.829 & -\newline -\newline - & -\newline -\newline - & -\newline -\newline - & -\newline -\newline - \\ \hline
Al-Smadi et al. (2018) \cite{alsmadi2018rnnvsvm} & Deep Learning (RNN) & Reviews (Hotels) \cite{pontiki2016semeval} & 0.870 & - & - & - & - \\ \hline
Yang et al. (2018) \cite{yang2018financial} & Deep Learning (RNN) & Social Media (Finance) \cite{10.1145/3184558.3192301} & - & - & - & - & MSE = 0.080 \\ \hline
Dong et al. (2014) \cite{dong2014adaptive} & Deep Learning (RecNN) & Social Media (Various) \cite{dong2014adaptive} \newline Reviews (Restaurants) \cite{pontiki-etal-2014-semeval} \newline (reimplemented in \cite{nguyen2015phrasernn}) & 0.663\newline 0.604 & -\newline 0.368 & -\newline 0.604 & 0.659\newline 0.457 & - \\ \hline
Nguyen and Shirai (2015) \cite{nguyen2015phrasernn} & Deep Learning (RecNN) & Reviews (Restaurants) \cite{pontiki-etal-2014-semeval} & 0.639 & 0.624 & 0.639 & 0.622 & - \\ \hline
Ruder et al. (2016) \cite{ruder2016insight} & Deep Learning (CNN) & Reviews (Electronics) \cite{pontiki2016semeval}\newline Reviews (Restaurants) \cite{pontiki2016semeval} \newline Reviews (Hotels) \cite{pontiki2016semeval}& 0.781 \newline 0.765 \newline 0.827 & -\newline -\newline - & -\newline -\newline - & -\newline -\newline - & -\newline -\newline - \\ \hline
Jangid et al. (2018) \cite{10.1145/3184558.3191827} & Deep Learning (CNN) & Social Media (Finance) \cite{10.1145/3184558.3192301} & - & - & - & - & MSE = 0.112 \\ \hline
Xue and Li (2018) \cite{xue-li-2018-aspect} & Deep Learning (CNN) & Reviews (Electronics) \cite{pontiki-etal-2014-semeval}\newline Reviews (Restaurants) \cite{pontiki-etal-2014-semeval} & 0.691 \newline 0.773 & -\newline - & -\newline - & -\newline - & -\newline - \\ \hline
Zeng et al. (2019) \cite{zeng2019aspect} & Deep Learning (CNN) & Reviews (Restaurants) \cite{pontiki-etal-2014-semeval} & 0.823 & - & - & - & - \\
\end{tabular}
}
\end{center}
\end{table*}

\subsubsection{Support Vector Machines}\label{sec:MethodsMachineLearningSVM}
Support vector machine (SVM) models \cite{cortes1995support} have long been a popular choice for sentiment analysis and ABSC \cite{alsmadi2018rnnvsvm, pang2002thumbs, pannala2016supervised, varghese2013svm, mullen2004sentiment}. SVM models are classifiers that distinguish between categories by constructing a hyperplane that separates data vectors belonging to the different classes \cite{cortes1995support, varghese2013svm, burges1998tutorial}. In the case of ABSC, this comes down to separating aspects into sentiment classes (\textit{``positive"}, \textit{``neutral"}, and \textit{``negative"}) based on the feature vector $\bm x$. Suppose we have a training dataset consisting of the $N$ feature vectors $\bm x_1, \dots, \bm x_N$ with the corresponding labels $y_1, \dots, y_N$. Each of the feature vectors $\bm x_1, \dots, \bm x_N$ represents an aspect with the corresponding context in a record, as explained in Section \ref{sec:Inputs}. The corresponding labels $y_1, \dots, y_N$ indicate the true sentiment expressed towards the aspects. We first examine the case when only two sentiment classes are considered: \textit{``positive"} and \textit{``negative"}. The labels for aspects for which a positive sentiment is expressed are encoded as a 1, while aspects toward which a negative sentiment is expressed receive a label of -1. The SVM classifier can then be summarized as follows:
\begin{equation}\label{equation:NonLinearSVM}
    \def\sss{\scriptscriptstyle}
    \setstackgap{L}{8pt}
    \def\stacktype{L}
    \stackunder{\hat{y}}{\sss 1 \times 1} = \text{sign}(\stackunder{\bm w^T}{\sss 1 \times d_z} \times \stackunder{\phi(\bm x)}{\sss d_z \times 1}+\stackunder{b}{\sss 1 \times 1}),
\end{equation}
where $\bm w \in \mathbb{R}^{d_x}$ is a learned vector of weights and $b \in \mathbb{R}^{1}$ is a learned bias constant. The weights are determined by constructing a hyperplane that maximizes the separation between the training data labels $y_1, \dots, y_N$ based on the feature vectors $\bm x_1, \dots, \bm x_N$. While some datasets can be separated using a linear function form, other problems may not be so easily separable. In such situations, the kernel function $\phi()$ can be used to transform the feature vector $\bm x$ into a higher dimension where the labels are separated more easily \cite{schlkopf2018learning}. The disadvantage of using a kernel function is that the learned coefficients become difficult to interpret due to the non-linearity. When an aspect can be classified into multiple sentiment categories, the SVM model must be adjusted. An example solution is a ``one-versus-all" implementation, where an SVM model is trained to separate each class versus all other classes. The final prediction is based on which decision function has the highest value. SVMs are known to generalize well and be robust towards noisy data \cite{xu2009robustness}. Furthermore, as mentioned in Section \ref{sec:Inputs}, feature vectors for ABSC can often consist of substantial amounts of features, which SVM models are known to handle well \cite{joachims1998text}. However, finding a kernel that works is generally a difficult task \cite{burges1998tutorial}, and hand-crafted features are required for the model to perform well \cite{alsmadi2018rnnvsvm}.

\subsubsection{Tree-Based}\label{sec:MethodsMachineLearningTree} Tree-based approaches are methods based on the trainable decision tree \cite{breiman1984classification} model. A decision tree model consists of a tree-like structure where each internal node, or decision node, in the tree represents a condition based on a particular feature from the vector $\bm x$, and each leaf node represents a particular sentiment class. Given the feature vector $\bm x$, we start at the root node of the tree and examine the splitting condition. The condition and the corresponding feature in $\bm x$ determine toward which internal node we move next. We move down the tree until a leaf node is reached, which has a certain sentiment class assigned to it, which determines the prediction $\hat{y}$. A significant advantage of decision trees is their explainability. The decision rules are typically easy to interpret for humans and can therefore be used to discover knowledge and obtain new insights  \cite{10.1007/978-3-030-65965-3_28}. The new insights can then also be used to, for example, improve the previously discussed knowledge bases. Although decision tree models have not been particularly popular for ABSC, there are some examples of works successfully implementing these models. In \cite{7972395}, an incremental decision tree model is proposed that outperforms an implementation of the previously discussed SVM model. Similarly, in \cite{akhtar2016aspect}, a decision tree is shown to outperform several other models, including an SVM, on a variety of datasets. However, SVM models can outperform decision tree models for other problems \cite{alsmadi2019enhancing}.

The main issue with decision trees is the problem of overfitting, which can be especially problematic for ABSC due to the substantial amounts of features that are often used. A solution to this problem is the use of a random forest model, which is an ensemble of decision trees \cite{breiman2001random}. A random forest consists of a large number of decision tree models. Each tree receives a limited number of features and a bootstrapped sample of the training data to train with. By randomly sampling data and restricting features, the individual decision trees tend to overfit less on specific features or data. Predictions are achieved via majority voting among the aggregated predictions of all individual decision tree models. In \cite{gupta-ekbal-2014-iitp}, a random forest model is implemented for the SemEval-2014 ABSC tasks, but mixed results are obtained for the different datasets. Similar results are presented in \cite{8126201}. Examples of other tree-based methods are the gradient boosted trees \cite{friedman2001greedy} and extra trees \cite{geurts2006extremely} classifiers. In \cite{bhoi2018various}, a comparison is made that indicates that these models may provide slight improvements compared to the random forest model.

\subsubsection{Deep Learning}\label{sec:MethodsMachineLearningDeep}
Deep learning models \cite{goodfellow2016deep} have revolutionized many fields of research \cite{sejnowski2018deep}, including sentiment analysis and ABSC \cite{zhou2019deep, do2019deep}. Significant amounts of research have been put into developing deep learning models for various types of data and learning tasks. One of the main disadvantages of deep learning models is the fact that they are highly difficult to interpret. Basic machine learning models, like decision trees and linear SVMs, can provide some useful model interpretations. Yet, although attempts have been made to explain the predictions produced by deep learning models \cite{guidotti2018survey, 10.1145/3359786}, these black-box methods are still regarded as practically not explainable. Furthermore, effectively training deep learning models requires large amounts of computational resources. This is because deep learning models require large amounts of data to train, which have not always been available for ABSC. However, as both deep learning research and the amount of publicly available data grow \cite{pontiki2016semeval, pontiki2015semeval, pontiki-etal-2014-semeval}, deep learning models become more and more popular due to their great predictive performances. Deep learning models used for ABSC include, but are not limited to, recurrent neural networks, recursive neural networks, and convolutional neural networks.

\begin{figure*}
    \centering
    \includegraphics[scale=0.6]{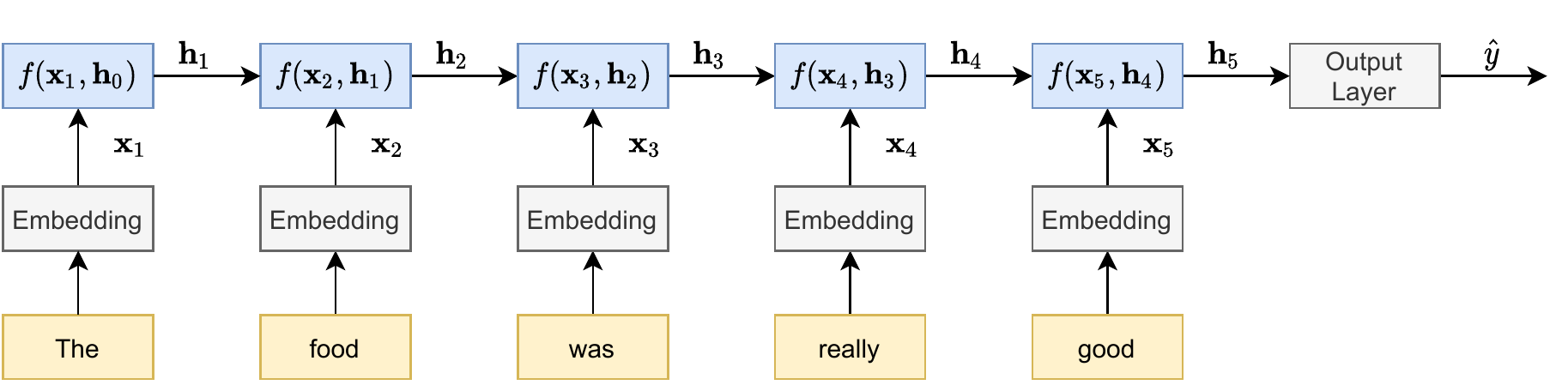}
    \caption{An illustration of a basic RNN model for ABSC.}
    \label{fig:RNN}
\end{figure*}

\textbf{Recurrent neural network} (RNN) models \cite{hopfield1982neural} have in recent years become one of, if not the most popular choice for ABSC models. RNN models are powerful tools used for learning sequence-based data \cite{lipton2015critical}. They have achieved remarkable results for many language-based learning tasks, including ABSC, but also a wide variety of other sequence-based tasks. A basic RNN model is presented in Figure \ref{fig:RNN}. In language processing, the main idea behind RNN models is that a sequence of words is sequentially fed through a neural network model. The hidden state produced by the neural network based on one word is used as input for the neural network at the next step, such that information is carried through the sequence. This same concept can also be used to process sequences of images, times series, or other sequences. We consider again a record $R$ containing an aspect $a$ with a label $y$. However, compared to the previously discussed classification methods, we now consider the case where the numeric features are represented in a matrix $\bm X \in \mathbb{R}^{d_x \times n_x}$, where $d_x$ represents the number of features used, and $n_x$ indicates the number of words considered to be in the context. For any task, given an input matrix $\bm X$, a general RNN model can be defined as follows:
\begin{equation}\label{equation:RNN}
    \def\sss{\scriptscriptstyle}
    \setstackgap{L}{8pt}
    \def\stacktype{L}
    \stackunder{\bm h_t}{\sss d_h \times 1} = f(\stackunder{\bm h_{t-1}}{\sss d_h \times 1}, \stackunder{\bm x_t}{\sss d_x \times 1}),
\end{equation}
where $\bm h_t \in \mathbb{R}^{d_h}$ is the hidden state vector at step $t$, $d_h$ is the pre-defined dimension of the hidden state vectors, and $\bm x_t \in \mathbb{R}^{d_x}$ is the $t$th column of $\bm X$, for $t = 1, \dots, n_x$. In the most basic RNN form, the function $f(.)$ represents a concatenation of the two vectors $\bm h_t$ and $\bm x_t$ that are then fed through a basic neural network model consisting of linear transformations and non-linear activation functions. Thus, at each step $t$, information from the previous words (contained in $\bm h_{t-1}$) is combined with the information contained in the following word (contained in $\bm x_t$). The final hidden state vector $\bm h_{n_x}$ should then contain all information from the words corresponding to the context of the aspect, processed from left to right. The last hidden state can then be put through an output layer to produce a label prediction. A typical example of an output layer is a linear transformation and a softmax function:
\begin{equation}\label{equation:FinalLayer}
    \def\sss{\scriptscriptstyle}
    \setstackgap{L}{8pt}
    \def\stacktype{L}
    \stackunder{\bm s}{\sss d_y \times 1} = \text{softmax}(\stackunder{\bm W_{f}}{\sss d_y \times d_h}\times \stackunder{\bm h_{n_x}}{\sss d_h \times 1} + \stackunder{\bm b_{f}}{\sss d_y \times 1}),
\end{equation}
where $\bm W_f \in \mathbb{R}^{d_y \times d_h}$ and $\bm b_f \in \mathbb{R}^{d_y}$ are, respectively, the trainable weight matrix and bias vector of the final layer, and $\bm s \in \mathbb{R}^{d_y}$ contains the probabilities of being the correct label for each sentiment class. These probabilities can also be interpreted as sentiment scores. After applying the softmax function, a label prediction can be obtained by predicting the label with the highest sentiment score or probability of being correct. While basic RNN models work well with shorter sequences, problems occur when they are used to process longer and more complex sentences. Due to the multiplicative nature of neural networks, RNN models generally suffer heavily from the vanishing gradient problem, which in turn means that these models have trouble learning long-term dependencies \cite{hochreiter2001gradient, bengio1994learning}. As such, more advanced RNN models, such as long short-term memory (LSTM) \cite{hochreiter1997long} and gated recurrent unit (GRU) \cite{cho2014properties} models, improve the function $f(.)$ by incorporating a series of gates. These gates allow for the information to flow through the model without losing critical details. These types of RNN models have been proven to work for many different problems and have also become the standard when implementing RNNs for ABSC. Further improvements can be made to RNN models by not only processing the words from left to right but also from right to left. These bi-directional RNN (Bi-RNN) models reverse the flow of information, which allows information at each end of the sequence of words to be preserved.

RNN models are versatile models that can be applied to many tasks that involve sequence data, including ABSC. An implementation of an LSTM model for sentence-level ABSC is presented in \cite{alsmadi2018rnnvsvm} where it is compared to an SVM model for ABSC of hotel reviews. The SVM model significantly outperforms the RNN model, which the authors explain to be due to the rich hand-crafted feature vectors used to train the SVM model. In contrast, in \cite{ruder2016hierarchical}, a hierarchical LSTM model is implemented and compared to other deep learning architectures using a collection of SemEval-2016 \cite{pontiki2016semeval} datasets. This hierarchical LSTM model leverages document-level information to perform sentence-level ABSC. The authors show that this model achieves competitive results compared to the best models of the competition, even though no rich hand-crafted features were used. In \cite{tang-etal-2016-effective}, a different approach is proposed that can be used when processing explicit aspects. Two LSTM models are used to model the left and right parts of the context relative to the target. This approach improves upon the basic LSTM model but is still similar in performance compared to advanced SVM models with rich features and pooled word embeddings \cite{vo2015target}. Nevertheless, RNN models seem to generally outperform SVMs, especially for more recent tasks (e.g., the FiQA-2018 task \cite{10.1145/3184558.3192301}). In recent research, deep learning models like RNNs are quickly outpacing SVM models in terms of predictive performance. This is the case for many language processing tasks, which is in part due to an increase in available training data. See for example the GLUE \cite{wang-etal-2018-glue} and SuperGLUE \cite{NEURIPS2019_4496bf24} benchmarks, where all the top-performing models are deep learning approaches. Yet, as previously mentioned, the increase in performance comes at a great computational cost, since deep learning models typically require significantly more resources to train than simpler models like SVMs. On the other hand, time is saved on designing handcrafted features. Therefore, when determining the optimal model choice for a certain task, one must take into consideration various factors besides predictive performance on benchmark datasets. Namely, there may not be enough computational resources available, or it could be too costly to obtain enough training data to be able to properly train a deep learning model. On the other hand, there may not be enough time to design handcrafted features for a high-performance SVM.

\begin{figure}
    \centering
    \includegraphics[scale=0.6]{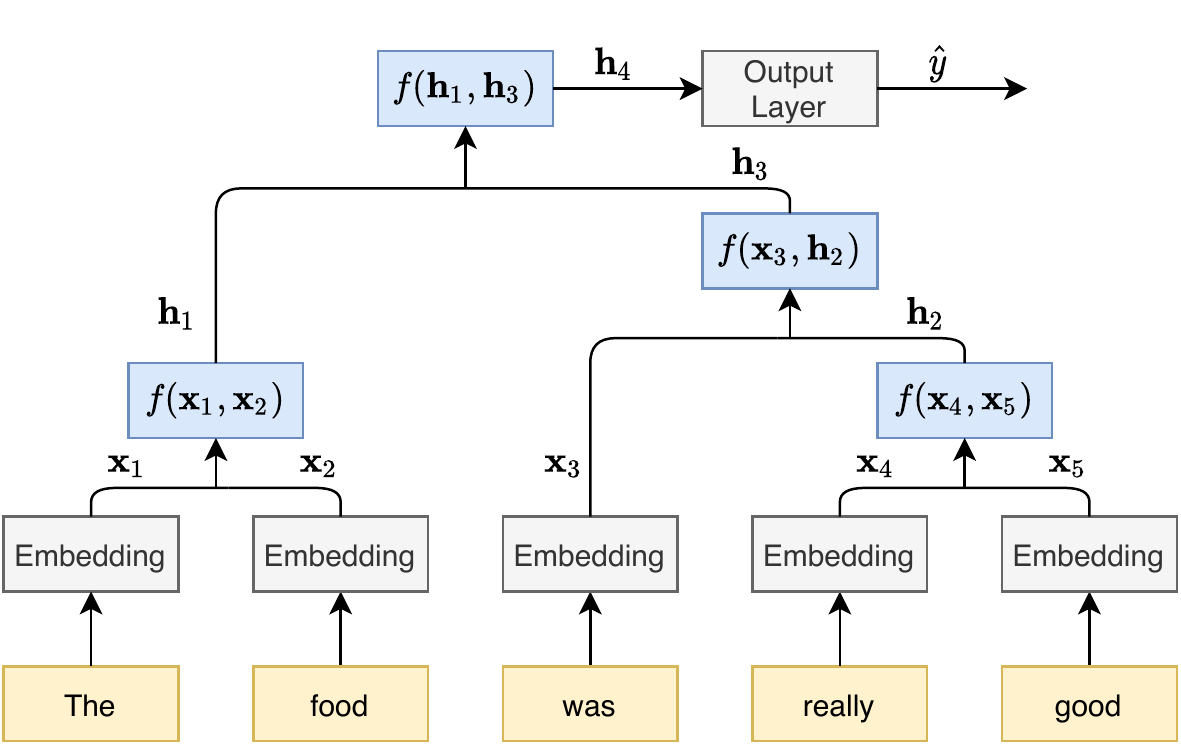}
    \caption{An illustration of a basic RecNN model for ABSC.}
    \label{fig:RecNN}
\end{figure}

\textbf{Recursive neural network} (RecNN) models \cite{goller1996learning} are a generalization of RNN models that process the words using a tree-like structure. Similar to RNNs, RecNNs are general models that can be applied to a variety of tasks. A basic RecNN model is presented in Figure \ref{fig:RecNN}. Similarly to RNN models, a function $f(.)$ is defined that is used to combine input vectors. This function is shared throughout the network and is therefore used in each processing step. As seen in Figure \ref{fig:RecNN}, RecNN trees consist of leaf nodes, which are the embedded input vectors, and internal nodes (depicted in blue) where vectors are combined using the function $f(.)$. Given a certain internal node with two child nodes, the output of the internal node can be defined as follows:
\begin{equation}\label{equation:RecNN}
    \def\sss{\scriptscriptstyle}
    \setstackgap{L}{8pt}
    \def\stacktype{L}
    \stackunder{\bm h}{\sss d_x \times 1} = f(\stackunder{\bm c_{1}}{\sss d_x \times 1}, \stackunder{\bm c_2}{\sss d_x \times 1}),
\end{equation}
where $\bm h \in \mathbb{R}^{d_x}$ is the output of the internal node, and $\bm c_1 \in \mathbb{R}^{d_x}$ and $\bm c_2 \in \mathbb{R}^{d_x}$ are the outputs of the child nodes. By iteratively combining the word embeddings and internal node output vectors, the model can extract information from the words. After the final step, the output vector $\bm h_{4}$ in Figure \ref{fig:RecNN} should contain the information from the sentence or context. While the example we used considers a binary tree, which is the most common type in ABSC, trees with three or more children per internal node are also possible. Although, the function $f(.)$ would have to be adjusted to process more than two inputs. In the most basic RecNN setup, the function $f(.)$ takes the form of a standard neural network. However, RecNN models using this basic setup may suffer from the same problems as RNNs do when using this function. As such, gated functions like the LSTM module have been adapted for RecNN models \cite{tai2015improved}.

The main advantage of using RecNN models over RNNs is the fact that the model is no longer restricted to processing the words sequentially from left to right and right to left. The order in which the inputs are processed depends on the type of relations that define the tree. The tree may also be structured such that the model processes the words in the original order, meaning that it would be equivalent to a standard RNN. However, herein also lies one of the disadvantages of these models. Namely, choosing how to define the tree structure can be difficult. RecNN trees are generally constructed using a language parser, which analyses the structure and dependencies of a sentence and deconstructs the sequence of words. Parsing is an entirely separate field in which much research has been done \cite{kubler2009dependency}. Various parsers exist, but the tree in Figure \ref{fig:RecNN} has been generated using the popular Stanford neural network parser \cite{chen2014fast}. 

The previously discussed RecNN characteristics are general attributes that apply to any task. Yet, more specialized RecNN models have been proposed for ABSC. For example, trees can be constructed in a manner that conforms specifically to the task of ABSC. In \cite{dong2014adaptive}, a technique is proposed that can be used to process explicit aspects. The trees in \cite{dong2014adaptive} are built towards the target word or phrase corresponding to the explicit aspect. This means that the model learns to forward-propagate the sentiment towards the aspect. Additionally, instead of using only one function $f(.)$, the proposed model implements multiple types of combination functions from which the model adaptively selects based on input vectors and linguistic characteristics. This RecNN specifically designed for ABSC is shown to significantly outperform previous SVM models \cite{jiang2011target}. Nguyen and Shirai \cite{nguyen2015phrasernn} expand on this idea by constructing a model that incorporates both dependency and constituent trees. Furthermore, they also expand upon the use of multiple combination functions. This model is shown to outperform the models presented in \cite{dong2014adaptive}. However, based on the reported results, the RecNN models we found still do not perform as well as other deep learning models.

\begin{figure}
    \centering
    \includegraphics[width=0.6\linewidth]{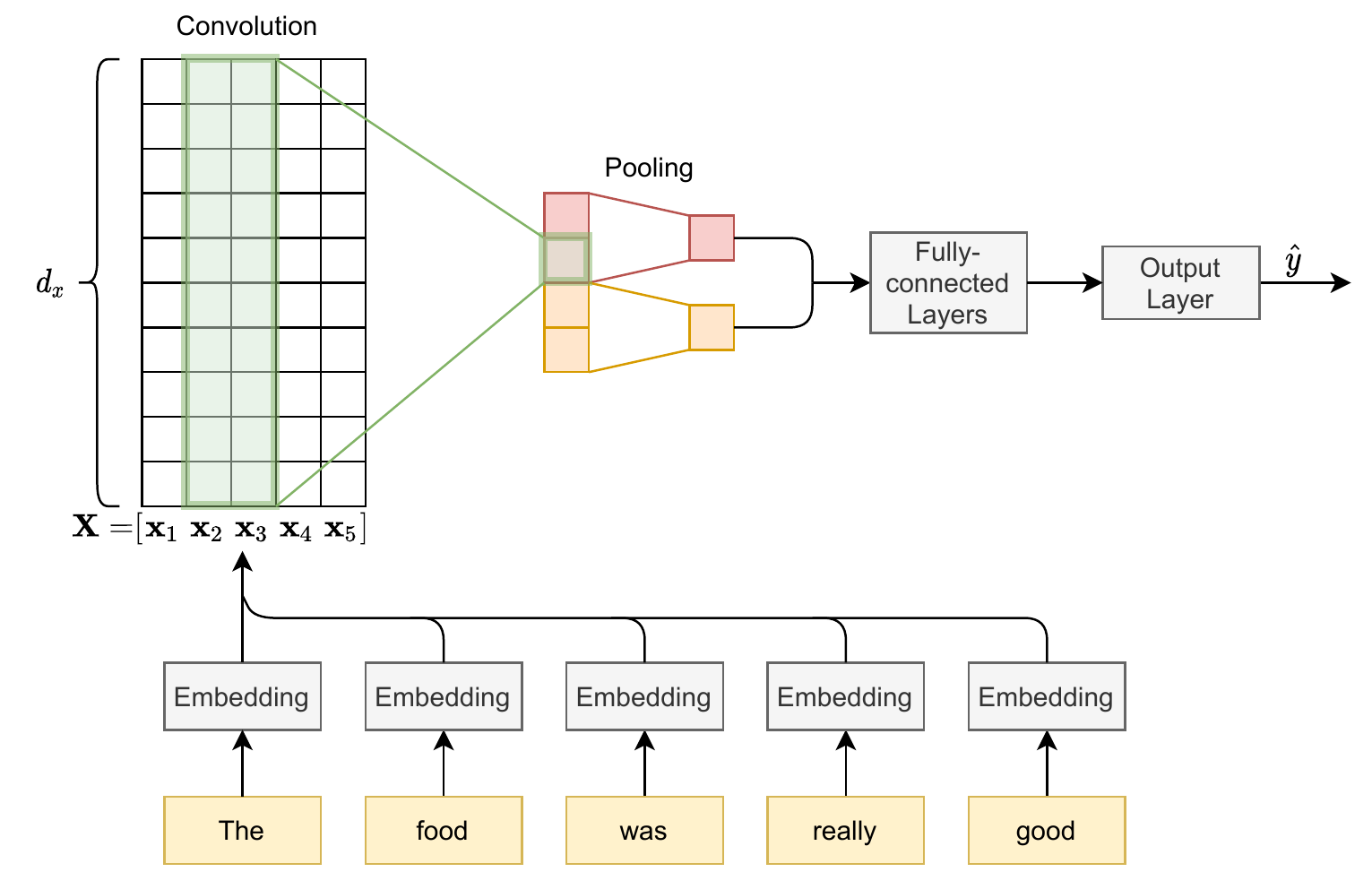}
    \caption{An illustration of a basic CNN model for ABSC. Each square box represents a value or element of a vector or matrix.}
    \label{fig:CNN}
\end{figure}

\textbf{Convolutional neural network} (CNN) models \cite{lecun1999object} are yet another popular deep learning model for text analysis methods, such as sentiment analysis and ABSA \cite{severyn2015twitter, xue-li-2018-aspect}. Originally, CNN models were used for processing images using three distinct layer types: convolutional layers, pooling layers, and fully-connected layers. A basic CNN model illustrating these layers is presented in Figure \ref{fig:CNN} in the context of language processing. The convolutional layer is generally described as a filter that slides across the input data matrix and produces linear combinations of the values within the window. Pooling layers can be used to reduce the dimensionality of the output and improve generalization by summarizing the information obtained from the convolutional layer. A global pooling layer is typically also used before the final layers to transform the features matrix into a single vector. This vector is then processed using fully-connected layers to produce the correct output.

The previously explained model layers form a general model architecture that can be used for many tasks. As such, CNN models have also been used for ABSC. In \cite{mulyo2018aspect}, CNN models are shown to produce superior performances compared to SVM models for ABSC. In \cite{ruder2016insight}, a CNN specifically designed for ABSC is implemented by including target embeddings so that the aspect can be modeled explicitly. The aspect embedding is concatenated with the input vectors to be used as input for the CNN. The reported results indicate that this approach allows for performances on par with the top models for the used datasets. A different approach presented in \cite{xue-li-2018-aspect} also uses aspect embeddings but in the form of a non-linear gating mechanism inserted between the convolutional and pooling layers. This model produces highly accurate predictions, outperforming many other models, including a random forest \cite{gupta-ekbal-2014-iitp}, RecNN models \cite{dong2014adaptive, nguyen2015phrasernn}, and even several attention-based deep learning models which are discussed in Subsection \ref{sec:MethodsMachineLearningAttention}. In \cite{zeng2019aspect}, this gated CNN is further expanded and improved using a linguistic regularization expansion of the loss function. Although this model comes close to being the best performing approach for the SemEval-2014 dataset, it was still outperformed by a hybrid model \cite{kiritchenko2014nrc}, which is discussed in Subsection \ref{sec:MethodsDictionaryEnhanced}.

\begin{figure*}
    \centering
    \includegraphics[scale=0.6]{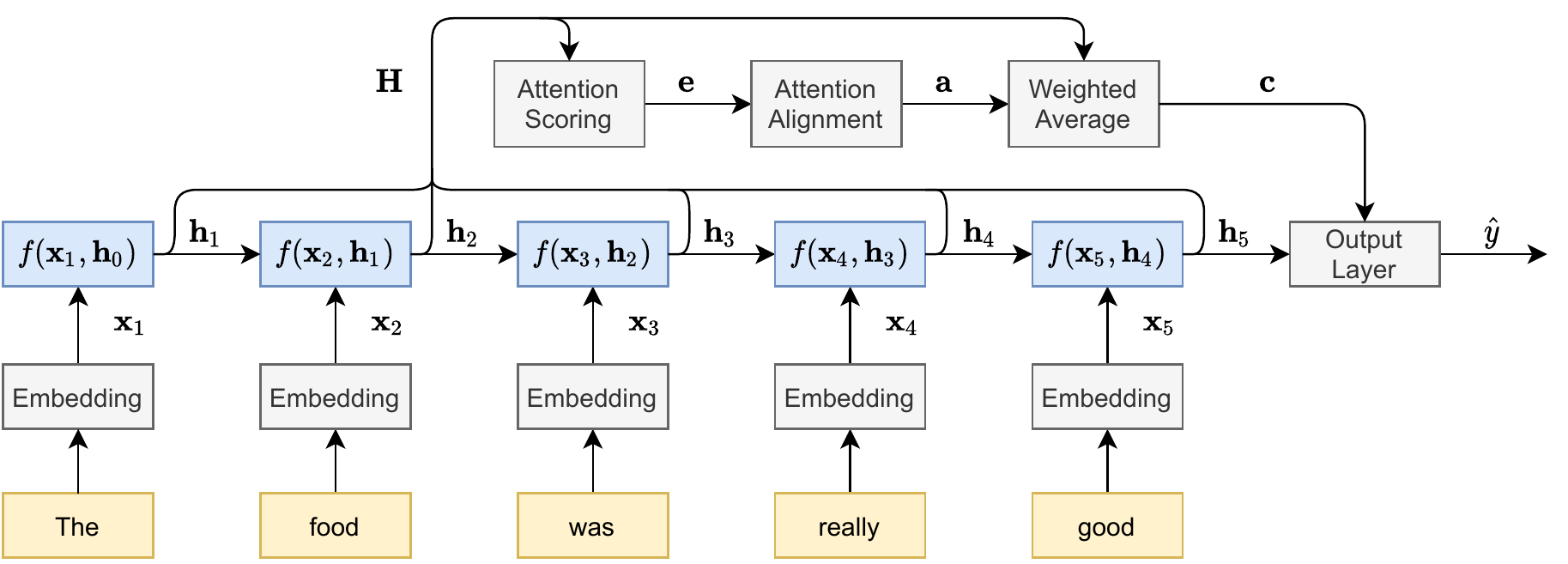}
    \caption{An illustration of a basic attention-based RNN model for ABSC.}
    \label{fig:AB-RNN}
\end{figure*}

\subsubsection{Attention-Based Deep Learning}\label{sec:MethodsMachineLearningAttention}
The attention mechanism \cite{bahdanau2014neural} is a highly effective extension to the deep learning models discussed in the previous subsection. in this section, we discuss attention-based deep learning (AB-DL) models for ABSC. The performances of various AB-DL models are displayed in Table \ref{Table:AB-DL}. We illustrate the attention mechanism using Figure \ref{fig:AB-RNN} in which the example RNN model presented in Figure \ref{fig:RNN} is extended with a basic attention architecture. Consider the input representation matrix $\bm X \in \mathbb{R}^{d_x \times n_x}$ for which the columns are the vectors $\bm x_1, \dots, \bm x_{n_x} \in \mathbb{R}^{d_x}$. RNN units are used to encode the sequence of words to produce the hidden states $\bm h_1, \dots, \bm h_{n_x} \in \mathbb{R}^{d_h}$ that correspond to the $n_x$ words. For clarity, we concatenate the hidden states to create the hidden state matrix $\bm H \in \mathbb{R}^{d_h \times n_x}$. Due to the sequential nature of the processing of the input vectors, the last hidden state $\bm h_{n_x}$ ($\bm h_{5}$ in Figure \ref{fig:AB-RNN}) should contain information from the entire sequence. However, as previously mentioned, RNN models struggle to learn long-term dependencies, which can be addressed using the attention mechanism.

\begin{table*}
\caption{Overview of prominent attention-based deep learning ABSC models and their reported performances.}

\label{Table:AB-DL}
\begin{center}
\resizebox{400pt}{!}{%
\begin{tabular}{p{125pt}|p{140pt}|p{135pt}|p{40pt}|p{40pt}|p{28pt}|p{20pt}}
\textbf{Reference} & \textbf{Attention Characteristics} & \textbf{Domain(s)} & \textbf{Accuracy} & \textbf{Precision} & \textbf{Recall} & \textbf{F$_1$} \\ \hline
Wang et al. (2016) \cite{wang2016attention} & Additive + Global + LSTM & Reviews (Electronics) \cite{pontiki-etal-2014-semeval}\newline Reviews (Restaurants) \cite{pontiki-etal-2014-semeval} & 0.687 \newline 0.772 & -\newline - & -\newline - & -\newline -  \\ \hline
Yang et al. (2017) \cite{yang2017attention} & General + Global + LSTM & Social Media (Various) \cite{dong2014adaptive} & 0.726 & - & - & 0.722  \\ \hline
Liu et al. (2018) \cite{liu2018cabasc} & Additive + Global + GRU & Reviews (Electronics) \cite{pontiki-etal-2014-semeval}\newline Reviews (Restaurants) \cite{pontiki-etal-2014-semeval}\newline Social Media (Various) \cite{dong2014adaptive}\newline Reviews (Restaurants) \cite{ pontiki2016semeval, pontiki2015semeval}\newline (reimplemented in \cite{wallaart2019hybrid})& 0.751 \newline 0.809\newline 0.715\newline 0.806 & -\newline -\newline -\newline - & -\newline -\newline -\newline - & -\newline -\newline -\newline - \\ \hline
Zhang et al. (2019) \cite{Zhang2019MultilayerAB} & Additive + Global + CNN & Reviews (Electronics) \cite{pontiki-etal-2014-semeval}\newline Reviews (Restaurants) \cite{pontiki-etal-2014-semeval}\newline Social Media (Various) \cite{dong2014adaptive}& 0.754 \newline 0.795\newline 0.713& -\newline -\newline - & -\newline -\newline - &-\newline -\newline - \\ \hline
Gan et al. (2020) \cite{GAN2020104827} & Additive + Global + CNN & Reviews (Electronics) \cite{pontiki-etal-2014-semeval}\newline Reviews (Restaurants) \cite{pontiki-etal-2014-semeval}\newline Social Media (Various) \cite{dong2014adaptive}& 0.736 \newline 0.814\newline 0.742& -\newline -\newline - & -\newline -\newline - &-\newline -\newline 0.732  \\ \hline
He et al. (2018) \cite{he2018effective} & Activated General + Syntax-Based + LSTM & Reviews (Electronics) \cite{pontiki-etal-2014-semeval}\newline Reviews (Restaurants) \cite{pontiki2016semeval, pontiki2015semeval, pontiki-etal-2014-semeval} & 0.719 \newline 0.823 & -\newline - & -\newline - & 0.692\newline 0.683  \\ \hline
Ma et al. (2017) \cite{ma2017ian} & Activated General + Global + Co-attention + LSTM & Reviews (Electronics) \cite{pontiki-etal-2014-semeval}\newline Reviews (Restaurants) \cite{pontiki-etal-2014-semeval}\newline Social Media (Various) \cite{dong2014adaptive}\newline (reimplemented in \cite{YANG2019463}) & 0.721 \newline 0.786 \newline 0.698 & -\newline -\newline -  & -\newline -\newline -  & -\newline -\newline -  \\ \hline
Gu et al. (2018) \cite{gu-etal-2018-position} & Activated General + Global + Co-attention + Bi-GRU & Reviews (Electronics) \cite{pontiki-etal-2014-semeval}\newline Reviews (Restaurants) \cite{pontiki-etal-2014-semeval} & 0.741 \newline 0.812 & -\newline - & -\newline - & -\newline - \\ \hline
Fan et al. (2018) \cite{fan2018multi}  & General + Global + Co-attention + Bi-LSTM & Reviews (Electronics) \cite{pontiki-etal-2014-semeval}\newline Reviews (Restaurants) \cite{pontiki-etal-2014-semeval}\newline Social Media (Various) \cite{dong2014adaptive}& 0.754 \newline 0.813\newline 0.725 & -\newline -\newline - & -\newline -\newline - & 0.725\newline 0.719\newline 0.708 \\ \hline
Yang et al. (2019) \cite{YANG2019463} & Additive + Global + Co-attention + LSTM & Reviews (Electronics) \cite{pontiki-etal-2014-semeval}\newline Reviews (Restaurants) \cite{pontiki-etal-2014-semeval}\newline Social Media (Various) \cite{dong2014adaptive}& 0.735 \newline 0.788\newline 0.715 & -\newline -\newline - & -\newline -\newline - & -\newline -\newline - \\ \hline
Zheng and Xia (2018) \cite{zheng2018left} & Activated General + Global + Rotatory + Bi-LSTM & Reviews (Electronics) \cite{pontiki-etal-2014-semeval}\newline Reviews (Restaurants) \cite{pontiki-etal-2014-semeval}\newline Social Media (Various) \cite{dong2014adaptive}\newline Reviews (Restaurants) \cite{ pontiki2016semeval, pontiki2015semeval} \newline (reimplemented in \cite{wallaart2019hybrid})& 0.752 \newline 0.813\newline 0.727\newline 0.827 & -\newline -\newline -\newline - & -\newline -\newline -\newline - & -\newline -\newline -\newline - \\ \hline
Tang et al. (2016) \cite{tang2016aspect} & Additive + Global + Multi-hop + LSTM & Reviews (Electronics) \cite{pontiki-etal-2014-semeval}\newline Reviews (Restaurants) \cite{pontiki-etal-2014-semeval}\newline Social Media (Various) \cite{dong2014adaptive}\newline (reimplemented in \cite{YANG2019463}) & 0.722 \newline 0.810 \newline 0.685 & -\newline -\newline -  & -\newline -\newline -  & -\newline -\newline -  \\ \hline
Fan et al. (2018) \cite{10.1145/3209978.3210115} & Additive + Multi-Hop + Position-Based + Bi-GRU & Reviews (Electronics) \cite{pontiki-etal-2014-semeval}\newline Reviews (Restaurants) \cite{pontiki-etal-2014-semeval}\newline Social Media (Various) \cite{dong2014adaptive} & 0.764 \newline 0.783 \newline 0.721 & -\newline -\newline - & -\newline -\newline - & 0.721\newline 0.684 \newline 0.708 \\ \hline
Majumder et al. (2018) \cite{majumder-etal-2018-iarm} & Additive/Multiplicative + Global + Multi-Hop + GRU & Reviews (Electronics) \cite{pontiki-etal-2014-semeval}\newline Reviews (Restaurants) \cite{pontiki-etal-2014-semeval} & 0.738 \newline 0.800 & -\newline - & -\newline - & -\newline - \\ \hline
Wallaart and Frasincar (2019) \cite{wallaart2019hybrid} & Activated General + Global + Rotatory + Multi-Hop + Bi-LSTM & Reviews (Restaurants) \cite{ pontiki2016semeval, pontiki2015semeval} & 0.831 & - & - & - \\ \hline
Gao et al. (2019) \cite{gao2019bert} & Transformer & Reviews (Electronics) \cite{pontiki-etal-2014-semeval}\newline Reviews (Restaurants) \cite{pontiki-etal-2014-semeval}\newline Social Media (Various) \cite{dong2014adaptive}& 0.784 \newline 0.846\newline 0.773 & -\newline -\newline - & -\newline -\newline - & 0.744\newline 0.796\newline 0.744  \\ \hline
Xu et al. (2019) \cite{xu-etal-2019-bert} & Transformer & Reviews (Electronics) \cite{pontiki-etal-2014-semeval}\newline Reviews (Restaurants) \cite{pontiki-etal-2014-semeval} & 0.781 \newline 0.850 & -\newline - & -\newline - & 0.751\newline 0.770  \\ \hline
Zeng et al. (2019) \cite{zeng2019lcf} & Transformer & Reviews (Electronics) \cite{pontiki-etal-2014-semeval}\newline Reviews (Restaurants) \cite{pontiki-etal-2014-semeval}\newline Social Media (Various) \cite{dong2014adaptive}& 0.825 \newline 0.871\newline 0.773 & -\newline -\newline - & -\newline -\newline - & 0.796\newline 0.817\newline 0.758 \\ \hline
Sun et al. (2019) \cite{sun2019bert} & Transformer & Social Media (Neighborhoods) \cite{saeidi-etal-2016-sentihood} & 0.933 & - & - & - \\ \hline
Karimi et al. (2020) \cite{Karimi2020AdversarialTF} & Transformer & Reviews (Electronics) \cite{pontiki-etal-2014-semeval}\newline Reviews (Restaurants) \cite{pontiki-etal-2014-semeval} & 0.794 \newline 0.860 & -\newline - & -\newline - & 0.765\newline 0.792 \\ \hline
Xu et al. (2020) \cite{XU2020135} & Transformer + Co-attention & Reviews (Electronics) \cite{pontiki-etal-2014-semeval}\newline Reviews (Restaurants) \cite{pontiki2016semeval, pontiki2015semeval, pontiki-etal-2014-semeval}\newline Social Media (Various) \cite{dong2014adaptive}& 0.781 \newline 0.843\newline 0.766 & -\newline -\newline - & -\newline -\newline - & 0.732\newline 0.712\newline 0.722 \\ \hline
Ansar et al. (2021) \cite{ansar2021efficient} & Transformer & Reviews (Movies) \cite{imdb, times}\newline Social Media (Neighborhoods) \cite{saeidi-etal-2016-sentihood} & 0.952 \newline 0.924 & -\newline - & -\newline - & -\newline - \\ \hline
Su et al. (2021) \cite{SU2021103477} & Transformer & Reviews (Electronics) \cite{pontiki-etal-2014-semeval}\newline Reviews (Restaurants) \cite{pontiki-etal-2014-semeval}\newline Social Media (Various) \cite{dong2014adaptive}& 0.826 \newline 0.879\newline 0.776 & -\newline -\newline - & -\newline -\newline - & 0.793\newline 0.823\newline 0.765 \\
\end{tabular}
}
\end{center}
\end{table*}

The basic attention architecture consists of three operations: attention scoring, attention alignment, and a weighted averaging operation. In the most general form, the attention mechanism requires three inputs: \textbf{key} vectors, \textbf{value} vectors, and a \textbf{query} vector. These attention concepts and the corresponding notation were introduced in \cite{daniluk2017frustratingly} and further popularized in \cite{vaswani2017attention}. The keys and values are generally derived from the data matrix on which attention is calculated. In the example presented in Figure \ref{fig:AB-RNN}, the data matrix upon which we calculate the attention is the matrix of hidden vectors $\bm H = [\bm h_1, \dots, \bm h_{n_x}]$. Key vectors can be derived by linearly transforming the hidden state vectors using a trainable weights matrix $\bm W_K \in \mathbb{R}^{d_k \times d_h}$, which produces the key vectors $\bm k_1, \dots, \bm k_{n_x} \in \mathbb{R}^{d_k}$. A similar process can be performed to obtain the value vectors $\bm v_1, \dots, \bm v_{n_x} \in \mathbb{R}^{d_v}$ using the trainable weights matrix $\bm W_V \in \mathbb{R}^{d_v \times d_h}$. However, for simplicity's sake, the example in Figure \ref{fig:AB-RNN} simply uses the original hidden state vectors as both the keys and values. The keys are used in combination with the query vector $\bm q \in \mathbb{R}^{d_q}$ in the first step of the attention calculation: attention scoring, which involves calculating an attention score corresponding to each feature vector. This score directly determines how much attention is focused on each word. The attention score for the $t$th word depends on the corresponding key vector $\bm k_t$, the query vector $\bm q$, and a score function. The score function can take a variety of forms, but it is generally meant to determine a relation between the key and query vectors. One of the most common score functions is the \textbf{additive} score function \cite{bahdanau2014neural}, which is used in many ABSC models \cite{wang2016attention}:
\begin{equation}\label{equation:AttentionScoring2}
    \def\sss{\scriptscriptstyle}
    \setstackgap{L}{8pt}
    \def\stacktype{L}
    \stackunder{e_t}{\sss 1 \times 1} = \stackunder{\bm w^T}{\sss 1 \times d_w}\hspace{-3pt} \times \text{act}(\stackunder{\bm W_k}{\sss d_w \times d_k} \hspace{-3pt}\times \hspace{-3pt}\stackunder{\bm k_t}{\sss d_k \times 1} \hspace{-3pt}+\hspace{-3pt} \stackunder{\bm W_q}{\sss d_w \times d_q}\hspace{-3pt}\times\hspace{-3pt} \stackunder{\bm q}{\sss d_q \times 1} \hspace{-3pt}+\hspace{-3pt} \stackunder{\bm b}{\sss d_w \times 1} \hspace{-5pt}),
\end{equation}
where $e_t \in \mathbb{R}^{1}$ is the attention score belonging to the $t$th word, $\bm w \in \mathbb{R}^{d_w}$, $\bm W_k \in \mathbb{R}^{d_w \times d_k}$, $\bm W_Q \in \mathbb{R}^{d_w \times d_q}$, and $\bm b \in \mathbb{R}^{d_w}$ are trainable weight matrices and vectors, and $d_w$ is a predefined dimension parameter. Other types of score functions are often based on a product, such as the  \textbf{multiplicative} \cite{luong2015effective}, \textbf{scaled-multiplicative} \cite{vaswani2017attention}, \textbf{general} score function \cite{luong2015effective}, and \textbf{activated general} \cite{ma2017ian} score functions. One can interpret attention scoring as determining which words contain the most important information with regards to determining the correct sentiment classification. The higher the score, the more important the information contained in the corresponding vector. Based on the scoring calculation, the query $\bm q$ is essential in determining what information is important. The simplest method of defining this query is by setting it as a constant vector. This allows the model to learn a general way of defining which words to focus on. However, since aspects can differ widely in their characteristics, a general query may not be flexible enough. A more popular technique is to use a vector representation of the aspect as the query. As seen in \cite{yang2017attention}, one can take the hidden vector representation of the target (a pooled version of $\bm h_1$ and $\bm h_2$ in Figure \ref{fig:AB-RNN}) as the query. In this case, the attention score calculation determines which information is most important with regard to the aspect itself, which enhances the model's ability to focus on the most important words in the sequence. This is especially important when a sentence contains multiple aspects. The purpose of the attention scores is to be used as weights in a weighted average calculation. However, the weights are required to add up to one for this purpose. As such, an alignment function align$()$ is used on the attention scores $e_1, \dots, e_{n_x} \in \mathbb{R}^{1}$:
\begin{equation}\label{equation:AttentionAlign}
    \def\sss{\scriptscriptstyle}
    \setstackgap{L}{8pt}
    \def\stacktype{L}
    \stackunder{a_t}{\sss 1 \times 1} = \text{align}(\stackunder{e_t}{\sss 1 \times 1}; \stackunder{\bm e}{\sss n_x \times 1}),
\end{equation}
where $a_t \in \mathbb{R}^{1}$ is the attention weight corresponding to the $t$th word, and we define the vector $\bm e \in \mathbb{R}^{n_x}$ as the vector containing all attention scores. A highly popular alignment function is known as \textbf{soft} or \textbf{global} alignment which applies a softmax function to the attention scores. Other types of alignment are, for example, the \textbf{hard} \cite{xu2015show} and \textbf{local} \cite{luong2015effective} alignment functions, which provide a more focused attention alignment. Additionally, in \cite{he2018effective}, the \textbf{syntax-based} alignment function is introduced that is specifically designed for ABSC. It employs global alignment but scales the attention weights according to how far the corresponding words are from the target in a dependency tree. The attention weights $a_1, \dots, a_{n_x} \in \mathbb{R}^{1}$ are used in combination with the value vectors $\bm v_1, \dots, \bm v_{n_x} \in \mathbb{R}^{d_v}$ to calculate a so-called \textbf{context} vector $\bm c \in \mathbb{R}^{d_v}$. For clarity, in Figure \ref{fig:AB-RNN} the attention weights are represented by the vector of attention weights $\bm a \in \mathbb{R}^{n_x}$. The context vector can be used in combination with the final hidden state in the output layer to obtain a label prediction. The context vector can be calculated as follows:
\begin{equation}\label{equation:ContextVector}
    \def\sss{\scriptscriptstyle}
    \setstackgap{L}{8pt}
    \def\stacktype{L}
    \stackunder{\bm c}{\sss d_v \times 1} = \sum^{n_x}_{t=1} \stackunder{a_{t}}{\sss 1 \times 1} \times \stackunder{\bm v_t}{\sss d_v \times 1}.
\end{equation}
This context vector summarizes the most important information contained in the context corresponding to the aspect. Since the attention mechanism can pick and choose information from any of the hidden states, information can be retrieved independently of where the it is positioned in the sequence. This significantly improves the model's ability to capture long-term dependencies. In addition, the attention weights are a direct indication of the importance of certain words. As such, attention models offer some explanation by analysing which words the model tends to focus on. This provides some interpretability for models which are typically considered black-box models. Although, the usefulness of the interpretation of attention weights is controversial \cite{jain2019attention, wiegreffe-pinter-2019-attention, mohankumar-etal-2020-towards}.

The example we used incorporates the hidden states from an RNN model into the attention calculation. However, attention can also be applied in other models such as a CNN. In the case of a CNN, the sets of feature maps that are obtained after the convolutional layer can be used as input for the attention mechanism. The rest of the attention calculation remains the same. The score and alignment functions are basic aspects that are required for every attention model. While the basic attention model we discussed can be used for the task of ABSC, extensions are often implemented for further improvements or to process other types of inputs. For example, \textbf{co-attention} \cite{NIPS2016_6202} can be used to calculate attention between multiple features matrices if there are multiple model inputs. This can be the case for ABSC if the input text is split up into parts, such as the target words and the words left and right of the target. A similar approach is called \textbf{rotatory} \cite{zheng2018left}, which rotates attention between the inputs. A more general attention extension is \textbf{multi-head} attention \cite{vaswani2017attention}, which employs so-called attention `heads' to calculate multiple different types of attention in parallel. In contrast, \textbf{multi-hop} attention \cite{tran2018multihop} is an extension that allows for multiple attention calculates in sequence. Multi-head and multi-hop attention are vital parts of the \text{transformer} model \cite{vaswani2017attention}. This model is based purely on the attention mechanism and does not use a separate base model like an RNN or CNN. This model uses \textbf{self-attention} to calculate attention between the feature vectors and use the extracted information to iteratively transform the input. Typical of this model type is the use of a scaled-multiplicative score function and a global alignment function since these functions are computationally highly efficient. The transformer model has been proven to be highly effective for many tasks, including ABSC. Gao et al. \cite{gao2019bert} present a transformer model with ABSC extensions for properly processing the target. A transformer model incorporating the previously mentioned co-attention mechanism is proposed in \cite{XU2020135}. Zeng et al. \cite{zeng2019lcf} implement two transformer models in parallel to allow the model to process the global context and local context concerning the aspect separately. All of these transformer models and extensions achieve significant improvements over previous models, including other attention models. When only considering predictive performance, transformer models tend to dominate most ABSC tasks and other modern language processing benchmarks, such as GLUE \cite{wang-etal-2018-glue} and SuperGLUE \cite{NEURIPS2019_4496bf24}. Yet, these models require large amounts of training data and computational resources to train from scratch, even compared to other deep learning models. A portion of current research is focused on alleviating the resources necessary for training transformer models via more efficient architectures, such as the linformer \cite{wang2020linformer}. Nevertheless, transformer models can produce impressive results, but may not be suitable for problems where the required resources are not available.

\subsection{Hybrid}\label{sec:MethodsHybrid}

When limited data is available, machine learning models may not provide satisfactory results \cite{severyn2015twitter}. ABSA is a domain where datasets are typically considered to be small in size \cite{he-etal-2018-exploiting} (see Table \ref{Table:Data}), which makes training large language models a significant challenge. This can be an even worse problem when attempting ABSC in some of its more niche sub-domains, such as sentiment analysis of texts for less popular products or topics. This problem is compounded by the fact that there is a vast number of different languages. English is a language for which there typically are extensive resources available. Yet, for other languages, like Arabic or Chinese, new datasets are slowly being introduced (see SemEval-2016 \cite{pontiki2016semeval}) but the amount of data is still limited \cite{do2019deep}. As such, large language models for ABSC can be hard to train. A solution to this problem is to incorporate additional knowledge from knowledge bases to compensate for the lack of data. We refer to models that incorporate both machine learning and knowledge bases as hybrid models. There are various ways in which knowledge bases can be combined with machine learning. A common approach is to use a knowledge base to define features that the machine learning algorithm uses to predict the sentiment. Another is to implement both a knowledge-based classifier and a machine learning classifier in sequence or parallel. In this section, we discuss these various approaches and categorize them using three categories: dictionary-enhanced, ontology-enhanced, and discourse-enhanced models. The performances of different hybrid models are presented in Table \ref{Table:Hybrid}.

\subsubsection{Dictionary-Enhanced Machine Learning}\label{sec:MethodsDictionaryEnhanced}

Dictionary-enhanced machine learning methods are some of the more common hybrid approaches. Dictionaries are rich knowledge bases that can easily be used to define features for a machine learning algorithm. In \cite{varghese2013svm}, sentiment scores from the \textit{SentiWordNet} \cite{baccianella2010sentiwordnet} dictionary are included as features to improve the performance of an SVM model. In both \cite{devi2016feature} and \cite{7975227}, the same technique using \textit{SentiWordNet} is also used. Additionally, \textit{NRC-Canada} \cite{kiritchenko2014nrc}, the top-performing model of the SemEval-2014 ABSC task, is based on a similar concept. However, the authors generate their own sentiment lexicons from unlabeled review corpora. In \cite{vo2015target}, a different approach is presented. First, word embeddings are defined using \textit{word2vec} \cite{mikolov2017advances}. Then, embeddings for words that are not in the sentiment lexicons are filtered out. As such, only words that contain useful sentiment information are kept. Since an SVM model is used for classification, the authors employ several pooling functions to summarize the features from the dense feature vectors to be used by the SVM model.

\begin{table*}
\caption{Overview of prominent hybrid ABSC models and their reported performances. ``AB-DL = Attention-Based Deep Learning"}

\label{Table:Hybrid}
\begin{center}
\resizebox{430pt}{!}{%
\begin{tabular}{p{115pt}|p{112pt}|p{130pt}|p{40pt}|p{40pt}|p{28pt}|p{20pt}}
\textbf{Reference} & \textbf{Category} & \textbf{Domain(s)} & \textbf{Accuracy} & \textbf{Precision} & \textbf{Recall} & \textbf{F$_1$} \\ \hline
Varghese and Jayasree (2013) \cite{varghese2013svm} & Dictionary-Enhanced SVM & Reviews (Movies) \cite{epinions}\newline Reviews (Music) \cite{pitchfork} & 0.853\newline 0.865 & -\newline - & -\newline - & -\newline - \\ \hline
Kiritchenko et al. (2014) \cite{kiritchenko2014nrc} & Dictionary-Enhanced SVM & Reviews (Electronics) \cite{pontiki-etal-2014-semeval}\newline Reviews (Restaurants) \cite{pontiki-etal-2014-semeval} & 0.705\newline 0.802 & -\newline - & -\newline - & -\newline -  \\ \hline
Vo et al. (2015) \cite{vo2015target} & Dictionary-Enhanced SVM & Social Media (Various) \cite{dong2014adaptive} & 0.711 & - & - & 0.699  \\ \hline
Devi et al. (2016) \cite{devi2016feature} & Dictionary-Enhanced SVM & Reviews (Electronics) \cite{amazon, ebay, flipkart} & 0.881 & 0.872 & 0.898 & 0.884  \\ \hline
Fachrina and Widyantoro (2017) \cite{8285850} & Dictionary-Enhanced SVM & Reviews (Various) \cite{8285850} & - & - & - & 0.858 \\ \hline
Akhtar et al. (2016) \cite{akhtar-etal-2016-hybrid} & Dictionary-Enhanced CNN & Reviews (Various) \cite{akhtar2016aspect_h} \newline Reviews (Electronics) \cite{pontiki-etal-2014-semeval}\newline Reviews (Restaurants) \cite{pontiki-etal-2014-semeval} & 0.660\newline 0.680\newline 0.772 & -\newline -\newline - & -\newline -\newline - & -\newline -\newline - \\ \hline
Ma et al. (2018) \cite{ma2018sentic} & Dictionary-Enhanced RNN & Reviews (Restaurants) \cite{pontiki2015semeval}\newline  Social Media (Neighborhoods) \cite{saeidi-etal-2016-sentihood} & 0.765 \newline 0.893 & -\newline - & -\newline - & -\newline -  \\ \hline
Ilmania et al. (2018) \cite{8629181} & Dictionary-Enhanced RNN & Reviews (Various) \cite{8285850} & - & - & - & 0.848  \\ \hline
Bao et al. (2019) \cite{bao-etal-2019-attention} & Dictionary-Enhanced AB-DL & Reviews (Restaurants) \cite{pontiki-etal-2014-semeval} & 0.829 & - & - & -  \\ \hline
Schouten et al. (2017) \cite{schouten2017ontology} & Ontology-Enhanced SVM & Reviews (Restaurants) \cite{pontiki2016semeval} & - & - & - & 0.753  \\ \hline
De Heij et al. (2017) \cite{10.1007/978-3-319-68786-5_27} & Ontology-Enhanced SVM & Reviews (Restaurants) \cite{pontiki2015semeval} & - & - & - & 0.808  \\ \hline
De Kok et al. (2018) \cite{de2018aggregated} & Ontology-Enhanced SVM & Reviews (Restaurants) \cite{pontiki2016semeval} & - & - & - & 0.812  \\ \hline
Schouten and Frasincar (2018) \cite{schouten2018ontology} & Ontology-Enhanced SVM & Reviews (Restaurants) \cite{ pontiki2016semeval, pontiki2015semeval} & 0.842 & - & - & -  \\ \hline
García-Díaz et al. (2020) \cite{GARCIADIAZ2020641} & Ontology-Enhanced RNN & Social Media (Diseases) \cite{GARCIADIAZ2020641} & 0.542 & - & - & - \\ \hline
Kumar et al. (2020) \cite{kumar2020aspect} & Ontology-Enhanced CNN & Reviews (Hotels) \cite{booking} & 0.885 & 0.943 & 0.856 & 0.860 \\ \hline
Me{\v{s}}kel{\.e} and Frasincar (2019) \cite{mevskele2019aldona} & Ontology-Enhanced AB-DL & Reviews (Restaurants) \cite{pontiki2016semeval}\newline  Reviews (Restaurants) \cite{ pontiki2016semeval, pontiki2015semeval} \newline (reimplemented in \cite{mevskele2020aldonar}) & 0.863 \newline 0.834 & -\newline - & -\newline - & -\newline -  \\ \hline
Wallaart and Frasincar (2019) \cite{wallaart2019hybrid} & Ontology-Enhanced AB-DL & Reviews (Restaurants) \cite{ pontiki2016semeval, pontiki2015semeval} & 0.843 & - & - & - \\ \hline
Trusca et al. (2020) \cite{trusca2020hybrid} & Ontology-Enhanced AB-DL & Reviews (Restaurants) \cite{ pontiki2016semeval, pontiki2015semeval} & 0.844 & - & - & - \\ \hline
Me{\v{s}}kel{\.e} and Frasincar (2020) \cite{mevskele2020aldonar} & Ontology-Enhanced AB-DL & Reviews (Restaurants) \cite{ pontiki2016semeval, pontiki2015semeval} &
0.855 & - & - & -  \\ \hline
Wang et al. (2018) \cite{ijcai2018-617} & Discourse-Enhanced AB-DL & Reviews (Electronics) \cite{pontiki2015semeval}\newline Reviews (Restaurants) \cite{pontiki2015semeval} & 0.816\newline 0.809 & -\newline - & -\newline - & 0.667\newline 0.685 \\ \hline
Wu et al. (2019) \cite{wu2019aspect} & Discourse-Enhanced AB-DL & Reviews (Electronics) \cite{pontiki-etal-2014-semeval}\newline Reviews (Restaurants) \cite{pontiki2016semeval, pontiki2015semeval, pontiki-etal-2014-semeval}\newline Social Media (Various) \cite{dong2014adaptive} & 0.734\newline 0.828\newline 0.698 & -\newline -\newline - & -\newline -\newline - & 0.691\newline 0.642\newline 0.675
\end{tabular}
}
\end{center}
\end{table*}

In \cite{bao-etal-2019-attention}, it is argued that attention models typically overfit when trained using small datasets. As such, \cite{bao-etal-2019-attention} proposes an attention-based LSTM model that incorporates lexicon features to improve the flexibility and robustness of attention-based deep learning models when trained with insufficient data. In \cite{akhtar-etal-2016-hybrid}, a hybrid model is presented specifically for resource-poor languages, such as Hindi. The proposed model combines trained CNN features and lexicon features. It is shown to produce better performances than the tested baselines that do not utilize the knowledge bases. In \cite{8629181}, a Bi-GRU model is proposed for ABSC for reviews from an Indonesian online marketplace. It is shown that performance is improved by incorporating lexicon features into the model.

Lexicons and other knowledge bases can enhance machine learning models, but machine learning can also improve the production of knowledge bases. An important example of this is \textit{SenticNet} \cite{10.1145/3340531.3412003}, which is a knowledge base that uses linguistic patterns, first-order logic, and deep learning to discover relationships between entities, concepts, and primitives. Such ensembles of symbolic and subsymbolic tools can be highly useful for sentiment analysis and ABSA. For example, \cite{ma2018sentic} proposed the Sentic LSTM model, which is an attention-based LSTM that fully incorporates sentic knowledge into the deep learning architecture. It is shown that this ensemble application of symbolic and subsymbolic AI produces better results than symbolic and subsymbolic AI separately.

\subsubsection{Ontology-Enhanced Machine Learning}\label{sec:MethodsOntologyEnhanced}

Ontologies can be used both as a knowledge base for sentiment information and to define a structure between concepts. In \cite{de2018aggregated}, a restaurant-specific ontology is used to define features for a review-level ABSC SVM model. The authors use both the concepts and the sentiment information from the ontology to define features. Essentially, a ``bag-of-concepts'' is defined that includes a binary feature for each concept in the ontology. However, the value of a concept feature depends on whether it is related to the aspect itself, meaning that implicit information in the text can be encoded into the feature vector. Secondly, features are defined to count how often sentiment polarity words from the ontology occur in the text.

A second approach for ontology-enhanced models is via a two-step method that sequentially employs a knowledge-based method and a machine learning classifier. First, an ontology-based model is employed to attempt to classify the aspect. When a conflicting answer is found for the sentiment label or when there is no information available regarding the sentiment, a backup model is employed in the form of a machine learning classifier. In \cite{wallaart2019hybrid}, sentence-level ABSC is performed by implementing a multi-hop rotatory attention model as the backup algorithm after an ontology-based classification model. The results presented indicate that the two-step method performs better compared to either of the classifiers individually. In \cite{mevskele2019aldona}, this two-step approach is used with a lexicalised domain ontology followed by an attention model. The attention model combines a sentence-level content attention mechanism and a bidirectional context attention mechanism. In \cite{mevskele2020aldonar}, this model is further extended using \textit{BERT} \cite{devlin2018bert} word embeddings, additional regularization, and adjustments to the training process. In \cite{trusca2020hybrid}, the two-step method using an ontology and a multi-hop rotatory attention model is extended with a hierarchical attention mechanism and deep contextual word embeddings. In \cite{schouten2018ontology}, both the two-step method and ontology-based features are used for sentence-level ABSC. First, an ontology-based classifier is used to try to determine a sentiment classification for the aspect. If no definitive answer can be found, an SVM is employed that uses a bag-of-words feature vector enhanced with ontology sentiment features.

Ontology-enhanced methods can prove useful when training language models for ABSC in niche domains. In \cite{10.1007/978-3-319-68786-5_27}, an ontology-enhanced SVM model is tested on the SemEval-2016 dataset. It is shown that a model with ontology features attains a significantly higher $F_1$-score than models that do not have those features. Moreover, the model with ontology features can obtain equal performance with less than 60\% of the training data. In \cite{schouten2017ontology}, it is shown that an ontology-enhanced model can be highly robust to changes in dataset size. For the task of aspect extraction, the performance of the ontology-enhanced SVM model proposed in \cite{schouten2017ontology} drops by less than 10\% with only 20\% of the original training data, while the performances of the base methods drop significantly. For the task of ABSC, all models proposed in \cite{schouten2017ontology} appear to be robust to dataset size changes. Yet, this can be explained by the fact that all methods are also enhanced using another knowledge base in the form of sentiment dictionaries. In \cite{GARCIADIAZ2020641}, an ontology-enhanced ABSC method is employed in the niche domain of infodemiology in the Spanish language. Aspects are extracted from Twitter posts and are classified using a Bi-LSTM model and an ontology of the infectious disease domain.

\subsubsection{Discourse-Enhanced Machine Learning}\label{sec:MethodsDiscourseEnhanced}

As discussed previously, discourse trees function similarly to ontologies in the sense that they do not inherently contain sentiment information. However, similarly to ontologies, the structure of discourse trees can still be useful to combine with a machine learning classifier. Wang et al. \cite{ijcai2018-617} use the structure of a discourse tree to determine how words are processed via an attention-based deep learning model for sentence-level ABSC. For each clause, a Bi-LSTM and an attention layer are used to produce a clause vector represented by the context vector output of the attention layer. The clause representations are then processed through another Bi-LSTM and attention layer to create a hierarchical attention structure. The resulting context vector is used for prediction. While the individual clauses defined by the discourse tree are incorporated in this model, the relational structure of the tree is not. In \cite{wu2019aspect}, the relations between discourse clauses are incorporated through the use of conjunction rules. First, each clause identified by the discourse tree is processed separately using Bi-LSTM layers. However, clause representations are produced simply by averaging. These clause representations are then used in the output layer, a layer to extract the relations between clauses using a Bi-LSTM, and a layer where conjunction rules are used to extract additional features. Conjunctions indicate how clauses are connected. Clauses connected by words such as \textit{``and"} are called coordinate conjunctions and often indicate a shared sentiment between the clauses. When words such as \textit{``but"} are used, an adversative conjunction is present that generally indicates an opposite sentiment. This information is used to selectively summarize the clause representations. Using these techniques, the rhetorical structure of the context can be incorporated into the model.

\section{Related Topics}\label{sec:Related}

\subsection{Aspect Detection \& Aggregation}\label{sec:RelatedDet&Agg}

In Section 1, ABSA was explained to consist of three steps: detection, classification, and aggregation. This survey discusses the classification step, also known as ABSC. Yet, the three steps cannot be completely separated from each other since these issues are not independent \cite{schouten2015survey}. For example, it may be important to consider the design of the classification and aggregation steps jointly, since information extracted during the classification step can be useful during the aggregation step. Namely, one can use features such as sentence importance to create a weighted average, which typically outperforms simple averages for aggregation \cite{basiri2020effect}. Similarly, information from the aspect detection step can be useful in the classification step. For example, one can use features extracted during the aspect extraction phase to predict the sentiments \cite{brun-etal-2014-xrce}. Yet, models have also been proposed that fully jointly perform the detection and classification steps. For example, in \cite{zhuang2006movie} a dictionary-based model using \textit{WordNet} \cite{miller1995wordnet} is proposed that extracts and pairs aspects and opinion words to find aspect-level sentiments. In \cite{schmitt-etal-2018-joint}, the detection and classification steps are completely fused in a single end-to-end LSTM model. The proposed model jointly learns the two tasks and produces improved performance for both tasks. In \cite{10.1145/3308558.3313750}, a capsule attention model is proposed that also jointly learns the detection and classification tasks in an end-to-end manner. This model produces state-of-the-art results for both tasks.

\subsection{Sarcasm \& Thwarting}\label{sec:RelatedSarcasmThwarting}

Thwarting and sarcasm are two complex linguistic phenomena that pose significant challenges for sentiment analysis models. Thwarting is the concept of building expectations, only to then contrast them. In other words, the overall sentiment of a document differs from the sentiment expressed throughout the majority of the text \cite{ramteke-etal-2013-detecting}. As such, methods that simply rely on aggregating the sentiment expressed by individual words can easily miss the greater context of the thwarted text. Additionally, developing a model for detecting thwarting is a difficult task due to the lack of available training data \cite{ramteke-etal-2013-detecting}. In the limit, detecting thwarting can also be considered to be highly similar to the difficult task of recognizing sarcasm \cite{ramteke-etal-2013-detecting}. Sarcasm is the act of deliberately ridiculing or questioning subjects by using language that is counter to its meaning \cite{joshi2018investigations}. It is particularly common in social media texts, where sarcasm analysis often also involves emoticons and hashtags \cite{maynard-greenwood-2014-cares}. Recognizing sarcasm in a text is a task that is often even difficult for humans \cite{maynard-greenwood-2014-cares} and requires significant world knowledge, which is difficult to include in most sentiment models \cite{ramteke-etal-2013-detecting}. 

The two problems of thwarting and sarcasm make sentiment analysis non-trivial \cite{annett2008comparison}. Yet, they have received little attention in the literature on ABSC. For ABSC, thwarting or sarcasm means that the sentiment expressed towards the aspect via the language used throughout the majority or the entirety of the record is opposite from the true sentiment. In the literature on general sentiment analysis, some attempts to address these problems have been proposed. Although, sentiment analysis is often considered a separate task from thwarting or sarcasm detection. For example, in \cite{ramteke-etal-2013-detecting}, a domain-ontology of camera reviews is used in combination with an SVM model to identify thwarted reviews of products. In \cite{8949523}, a multi-head attention-based LSTM model is used to detect sarcasm in social media texts. Yet, thwarting and sarcasm can also be handled in the context of sentiment analysis. For example, in \cite{mishra-etal-2016-leveraging}, the problems of thwarting and sarcasm in sentiment analysis are addressed using cognitive features obtained by tracking the eye movements of human annotators. It is theorized that the cognitive processes used to recognize thwarting and sarcasm are related to the eye movements of readers. It is shown that the gaze features significantly improve the performance of sentiment analysis classifiers, such as an SVM. Furthermore, it is specifically shown that the gaze features help address the problem of analysing complex linguistic structures, such as thwarting and sarcasm. In \cite{el-mahdaouy-etal-2021-deep}, the tasks of sentiment analysis and sarcasm detection are considered jointly via a multi-task model. A transformer model is used to identify both sentiment and sarcasm in Arabic tweets. It is shown that the proposed multi-task model outperforms its single-task counterparts. Such methods could also be used for ABSC. Cognitive features could be combined with aspect features to address thwarting and sarcasm for the task of ABSC. Similarly, the task of ABSC can be considered jointly with the task of sarcasm detection. Yet, more specialized ABSC datasets would be required for such problems.

\subsection{Emotions}\label{sec:RelatedEmotions}

Emotions are an interesting subject in conjunction with ABSC. Sentiment analysis and emotion analysis are two highly related topics. In sentiment analysis, we typically assign polarity labels or scores to texts, but in emotion analysis, a wide range of emotions (e.g., ``\textit{joy}'', ``\textit{sadness}'', ``\textit{anger}'') are to be considered \cite{10.1109/MIS.2016.31}. For example, in \cite{topal2016movie}, movie reviews are analysed and assigned emotion scores based on the dimensions of the hourglass of the emotions \cite{cambria2012hourglass}. Like sentiment analysis, emotion analysis can be done at multiple levels \cite{hakak2017emotion}. The equivalent of ABSC for emotion analysis is called aspect-based emotion analysis. For example, in \cite{polignano2018emotion}, emotions towards aspects in social media posts are classified using word embedding centroids, lexicons, and emoticons. More extensive work on the usage of emoticons in sentiment analysis is available from \cite{10.1145/2480362.2480498, hogenboom2015exploiting}. In \cite{suciati2020aspect}, several models are tested to detect emotions towards aspects in restaurant reviews. Information about emotions can also help with the task of ABSC. For example, in \cite{kumar2021sentic}, features from the \textit{SenticNet} \cite{10.1145/3340531.3412003} knowledge base are used to enhance a model for ABSA. In \cite{ma2018sentic}, an attention-based LSTM model is proposed that incorporates emotion information from the \textit{AffectiveSpace} knowledge base \cite{cambria2015affectivespace} for improved performance in ABSC tasks. A similar technique is used in \cite{ma2018targeted}. Additionally, emotion analysis and sentiment analysis can also be performed jointly. For example, in \cite{7933922}, a model is proposed that can extract aspect-level affective knowledge that includes sentiment polarities and emotion categories. In \cite{wang2020multi}, an extensive framework for multi-level sensing of sentiments and emotions is proposed that incorporates knowledge bases and sarcasm handling.

\section{Conclusion}\label{sec:Conclusion}

In this survey, we have provided an overview of the current state-of-the-art models for ABSC. We have explained the process of ABSC according to its three main phases: input representation, aspect sentiment classification, and output evaluation. The input representation phase involves representing a body of text by a numeric vector or matrix such that a classification model can identify the correct polarity label of an aspect. In the output evaluation phase, the quality of these polarity label predictions is assessed using performance measures. The quality of the predictions is determined by the architecture of the classification model. We have discussed a variety of state-of-the-art ABSC models using a proposed taxonomy and summarizing tables that serve as overviews of model performances. These ABSC models have been discussed and compared using intuitive explanations, technical details, and reported performances. We have also discussed a variety of important topics related to ABSC.

ABSC is a relatively new task that has quickly gained popularity and is rapidly changing. A noticeable evolution in the field of ABSC concerns the used datasets. The authors of the early ABSC works often scraped and compiled their own datasets from the Web. While this phenomenon makes sense due to the vast amounts of public reviews available online, it makes performance comparisons difficult due to most models being tested on different datasets. It was only after researchers started adopting the Twitter dataset from \cite{dong2014adaptive} and the review datasets from the SemEval challenge \cite{pontiki2016semeval, pontiki2015semeval, pontiki-etal-2014-semeval} that actual comparative analyses became more feasible. While this is a significant development for the domain of ABSC, a consequence is that ABSC models are mostly only implemented for restaurant reviews, electronics reviews, and Twitter data. Other types of data, such as hotel reviews, go mostly ignored even though the SemEval-2016 datasets \cite{pontiki2016semeval} contain a set of hotel reviews. To further advance the field of ABSC, it is desirable to test models in other domains. This requires more high-quality public datasets to be made available and adopted. Examples are the FiQA-2018 \cite{10.1145/3184558.3192301} and SentiHood \cite{saeidi-etal-2016-sentihood} datasets. Another example is the book review dataset produced by {\'A}lvarez-L{\'o}pez et al. \cite{alvarez2017book}. This dataset contains a subset of book reviews taken from the INEX Amazon/LibraryThing book corpus \cite{koolen2016overview} that was hand-annotated at the aspect level. Yet, this dataset has rarely been adopted by other researchers.

New datasets are also required for languages other than English. As discussed, English is a language for which there is a relatively sizeable number of datasets available for the task of ABSC. This is not the case for most other languages, which makes training language models a difficult task. If it is too costly to obtain more training data, simpler models like SVMs are typically the better option. However, the incorporation of knowledge bases may help compensate for the lack of data, as evidenced by the hybrid models discussed in Subsection \ref{sec:MethodsHybrid}. As has been shown in various works \cite{ijcai2018-617, mevskele2019aldona, trusca2020hybrid}, knowledge bases are effective in enhancing state-of-the-art models to achieve higher performances. We believe that the further exploration of knowledge-enhanced methods will help improve on the current state-of-the-art. However, the incorporation in current research is mostly relatively basic. For example, in \cite{ijcai2018-617}, only the clauses extracted by the discourse tree are incorporated in the model. The model presented in \cite{wu2019aspect} improves upon this by including conjunction rules, but this solution fails to exploit the many other discourse relations that can contain useful information for sentiment classification. Similarly, we have mainly seen domain ontologies in hybrid models as part of a two-step method where the ontology is separate from the machine learning model. Yet, ontologies contain many concepts and relations that can provide important features for the machine learning model. As such, we would advocate for further incorporation of knowledge bases and their structures. Furthermore, while we have only discussed three types of knowledge bases (dictionaries, ontologies, and discourse trees), we expect to see further exploration of new knowledge bases to be incorporated for ABSC. Additionally, improvements can be made in the construction of knowledge bases. This is exemplified by the recent research on the semi-automatic construction of domain ontologies for ABSC \cite{schouten2018ontology, zhuang2019soba, dera2020sasobus, tenhaaf2021websoba}. Another problem is that proper knowledge bases for resource-poor languages may be scarce as well \cite{akhtar2016aspect_h}. As such, the development of new knowledge resources other than labeled ABSC datasets is an important next step.

Often, the problem is not a lack of data, but specifically a lack of labeled data. Unsupervised or weakly supervised methods can be highly useful in situations where labeled data is too expensive to obtain. For example, in \cite{10.1145/3397271.3401179}, a weakly supervised model is proposed for joint aspect extraction and sentiment classification. The proposed model involves a sentiment dictionary learned via an auto-encoder using attention. Furthermore, weakly supervised systems can even be used to produce new training data via labeling mechanisms based on expert knowledge \cite{ratner2017snorkel}. An alternative solution to the problem of a lack of data is the use of cross-lingual and multi-lingual models. Knowledge from a language with extensive resources available can be transferred to models for other languages to compensate for the lack of data \cite{barnes-etal-2016-exploring}. For example, in \cite{akhtar-etal-2018-solving}, an attempt is made to solve the problem of data sparsity in French and Hindi datasets via the use of a deep learning model built on top of bilingual word embeddings. These bilingual word embeddings were produced using English-French and English-Hindi parallel corpora created via standard machine translation methods. Training models that are more language-agnostic is an important step for language models in general. Another important step is the development of domain-agnostic models. The problem of data scarcity for ABSC can be alleviated by transferring knowledge from other language domains where data is more readily available in large quantities \cite{xu-etal-2019-bert, sun2019bert}. Pre-training large language models like BERT on large language datasets from other domains and then fine-tuning the model parameters using a small domain dataset is a popular technique to handle this problem and can become more and more useful as larger and more general language models continue to emerge \cite{brown2020language}.

Early ABSC approaches were systems based almost purely on knowledge bases. As larger labeled datasets became available, machine learning models such as SVMs became the standard for ABSC. Soon after, deep learning methods started becoming more popular, but often performed on par with the machine learning methods that used high-quality handcrafted features. Yet, with the introduction of attention, deep learning methods started rapidly outpacing other approaches. Until a new revolutionary innovation comes along, we foresee attention-based deep learning models to be the future for ABSC. New attention models are rapidly being developed in and outside the field of ABSC. For example, multi-dimensional attention \cite{shen2018disan} is a general extension of the attention mechanism that allows for a more fine-grained attention computation. Yet, it has barely been explored for ABSA. Similarly, while multi-head attention is a general attention extension, it is typically only used in transformer-based architectures \cite{vaswani2017attention}. Admittedly, the transformer model is a highly successful model producing state-of-the-art performances in ABSC  \cite{gao2019bert, xu-etal-2019-bert, Karimi2020AdversarialTF, XU2020135, zeng2019lcf}. Transformer models will undoubtedly remain relevant as new transformer-based architectures are proposed that can be used for ABSC, such as the \textit{transformer-XL} \cite{DBLP:conf/acl/DaiYYCLS19} and the \textit{reformer} model \cite{kitaev2020reformer}. Attention models provide an inherent type of interpretability via the attention weights, which is an important aspect that is missing from many black-box algorithms like deep learning models. Another example of more explainable models is the ensembles of symbolic and subsymbolic AI. Models like the Sentic LSTM \cite{ma2018sentic} may be the start for new explainable models for ABSA \cite{susanto2021ten}.

Not only can the field of ABSC evolve through advancements in models and datasets, but also via research on the applications of ABSC. For example, most ABSC methods focus either on implicit or explicit aspects. Yet, texts can often contain multiple implicit and explicit aspects \cite{dosoula2016sentiment}. This poses a significant challenge for real-words applications of ABSA methods. As such, further research on methods that can handle both implicit and explicit aspects is required. Furthermore, when considering real-world applications of ABSA, users are typically interested in the sentiment expressed towards aspects aggregated over a review or sets of reviews. Yet, most ABSC methods are focused on sentence-level ABSC, after which the sentiment is aggregated at the review level. However, it has been shown that pure review-level ABSC can outperform sentence-level ABSC with aggregation \cite{de2018aggregated}. As such, more research on the application of review-level ABSC can also help move the field forward. An interesting new application for ABSC is the implementation of ABSA-based search engines. \textit{Smith} \cite{choi2015smith} is a specialized opinion-based search engine that returns restaurants based on sentiments expressed towards certain aspects in online reviews. Such specialized search engines and other new applications will drive the field of ABSC forward in the future. Another direction that is to be explored stems from the manner in which the task of ABSC is formulated. Currently, the task of ABSC is generally defined in the absence of a time element, meaning that sentiments are considered to be static over time. Although, sentiments towards products are known to typically change over time \cite{moe2012online}. This concept has been considered in some sentiment analysis research. For example, in \cite{wang2012system}, the overall sentiment expressed toward the presidential candidates in the 2012 U.S. election is tracked over time based on Twitter posts. Nonetheless, ABSA over time is an interesting problem that has not seen much attention. However, this development requires new datasets to be produced that are suitable for this task.

\bibliographystyle{ACM-Reference-Format}
\bibliography{references.bib}

\end{document}